\documentclass[twocolumn, switch, table, xcdraw]{article} 

\usepackage{preprint}
\usepackage{changepage}  

\usepackage{amsmath, amsthm, amssymb, amsfonts}
\usepackage{pdflscape}

\usepackage[numbers,square]{natbib}

\usepackage[utf8]{inputenc}	
\usepackage[T1]{fontenc}	
\usepackage{xcolor}		
\usepackage[colorlinks = true,
            linkcolor = purple,
            urlcolor  = blue,
            citecolor = cyan,
            anchorcolor = black]{hyperref}	
\usepackage{nicefrac}		
\usepackage{microtype}		
\usepackage{lineno}		
\usepackage{float}			
\usepackage{tabularray}
\UseTblrLibrary{functional}

\usepackage{pgf} 
\definecolor{low}{HTML}{00994d} 
\definecolor{high}{HTML}{fff51b}  
\pgfmathtruncatemacro{\opacity}{70} 
\pgfmathsetmacro{\minvall}{0}
\pgfmathsetmacro{\maxvall}{0.64}
\IgnoreSpacesOn
    \prgNewFunction \grhighlowColor {} {
        \intStepOneInline {2} {\arabic{rowcount}} {
            \intSet \lTmpaInt { \intMathMod {##1} {2} }
                \intStepOneInline {2} {\arabic{colcount}} {
                    \tlSet \lTmpbTl {\cellGetText {##1} {####1}}
                    \fpCompareTF {\lTmpbTl} > {\maxvall} { } {
                        \fpCompareTF {\lTmpbTl} < {\minvall} { } {
                            \pgfmathparse{int(round(100*(\lTmpbTl/(\maxvall-\minvall))-(\minvall*(100/(\maxvall-\minvall)))))}
                            \cellSetStyle {##1} {####1} {bg=high!\pgfmathresult!low!\opacity}
                        }
                    }
                }
        }
    }
\IgnoreSpacesOff

\usepackage{newfloat}
\DeclareFloatingEnvironment[name={Supplementary Figure}]{suppfigure}
\usepackage{sidecap}
\sidecaptionvpos{figure}{c}

\usepackage{titlesec}
\titlespacing\section{0pt}{12pt plus 3pt minus 3pt}{1pt plus 1pt minus 1pt}
\titlespacing\subsection{0pt}{10pt plus 3pt minus 3pt}{1pt plus 1pt minus 1pt}
\titlespacing\subsubsection{0pt}{8pt plus 3pt minus 3pt}{1pt plus 1pt minus 1pt}

\usepackage[toc,page]{appendix} 
\usepackage{subcaption} 
\usepackage{booktabs} 
\usepackage{dsfont} 
\usepackage{amssymb}
\usepackage{pifont}
\usepackage{stmaryrd} 
\usepackage{makecell} 
\usepackage{graphicx,multirow} 
\usepackage{colortbl} 
\usepackage{amsmath}
\usepackage{bm} 
\usepackage{url}  
\usepackage{siunitx} 
\usepackage{caption} 
\usepackage{float} 
\usepackage{threeparttable}
\usepackage{tabularx}

\usepackage{etoolbox}
\usepackage{tablefootnote}

\title{Cyto R-CNN and CytoNuke Dataset: Towards reliable whole-cell \mbox{segmentation} in bright-field histological images}

\usepackage{background}

\backgroundsetup{
    angle=0,
    scale=0.75,
    placement=bottom,
    color=black,
    vshift=20pt,
    contents={
        \ifnum\value{page}=1 
            *Correspondence to: Johannes Raufeisen, E-Mail: \tt{jraufeisen@ukaachen.de} $\dagger$ Behrus Puladi, E-Mail: \tt{bpuladi@ukaachen.de}          
        \else
            \ifnum\value{page}=12
                Preprint style based on the Latex Template (c) by Brenhin Keller, CC-BY-4.0, https://creativecommons.org/licenses/by/4.0/
            \fi
        \fi
    }
}

\usepackage{tikz,xcolor,hyperref}

\definecolor{lime}{HTML}{A6CE39}
\DeclareRobustCommand{\orcidicon}{
	\begin{tikzpicture}
	\draw[lime, fill=lime] (0,0) 
	circle [radius=0.16] 
	node[white] {{\fontfamily{qag}\selectfont \tiny ID}};
	\draw[white, fill=white] (-0.0625,0.095) 
	circle [radius=0.007];
	\end{tikzpicture}
	\hspace{-2mm}
}
\foreach \x in {A, ..., Z}{\expandafter\xdef\csname orcid\x\endcsname{\noexpand\href{https://orcid.org/\csname orcidauthor\x\endcsname}
			{\noexpand\orcidicon}}
}

\usepackage{authblk}

\author[1,2\thanks{}]{Johannes Raufeisen \orcidA}
\author[1,2]{Kunpeng Xie \orcidB}
\author[3,4]{Fabian Hörst \orcidC}
\author[6,7]{Till Braunschweig \orcidD}
\author[3,4]{Jianning Li \orcidE}
\author[3,4,5]{Jens Kleesiek \orcidF}
\author[2]{Rainer Röhrig \orcidG}
\author[3,4,8]{Jan Egger \orcidH}
\author[9]{Bastian Leibe \orcidI}
\author[1]{Frank Hölzle \orcidJ}
\author[9]{Alexander Hermans \orcidK}
\author[1,2\thanks{}]{Behrus Puladi \orcidL}

\affil[1]{\scriptsize Department of Oral and Maxillofacial Surgery, University Hospital RWTH Aachen, Pauwelsstr. 30, 52074 Aachen, Germany}
\affil[2]{Institute of Medical Informatics, University Hospital RWTH Aachen, Pauwelsstr. 30, 52074 Aachen, Germany }
\affil[3]{Institute for Artificial Intelligence in Medicine (IKIM), University Hospital Essen (AöR), Girardetstraße 2, 45131 Essen, Germany}
\affil[4]{Cancer Research Center Cologne Essen (CCCE), West German Cancer Center Essen, University Hospital Essen (AöR), Hufelandstr. 55, 45147 Essen, Germany}
\affil[5]{Department of Physics, TU Dortmund University, August-Schmidt-Str. 4, 44227 Dortmund, Germany}
\affil[6]{Institute of Pathology, University Hospital RWTH Aachen, Pauwelsstr. 30, 52074 Aachen, Germany }
\affil[7]{Institute of Pathology, Ludwig Maximilian University of Munich, Thalkirchner Str. 36, 80337 Munich, Germany }
\affil[8]{Center for Virtual and Extended Reality in Medicine (ZvRM), University Hospital Essen, University Medicine Essen, Hufelandstraße 55, 45147 Essen, Germany}
\affil[9]{Visual Computing Institute (Computer Vision), RWTH Aachen University, Mies-van-der-Rohe Str. 15, 52074 Aachen, Germany }

\normalsize

\begin{document}
\twocolumn[ 
  \begin{@twocolumnfalse} 
  
\maketitle

\begin{abstract}
    \textbf{Background:}
    Cell segmentation in bright-field histological slides is a crucial topic in medical image analysis. Having access to accurate segmentation allows researchers to examine the relationship between cellular morphology and clinical observations. Unfortunately, most segmentation methods known today are limited to nuclei and cannot segmentate the cytoplasm. 
    
    \textbf{Material \& Methods:}
    We present a new network architecture Cyto R-CNN that is able to accurately segment whole cells (with both the nucleus and the cytoplasm) in bright-field images. We also present a new dataset CytoNuke, consisting of multiple thousand manual annotations of head and neck squamous cell carcinoma cells. Utilizing this dataset, we compared the performance of Cyto R-CNN to other popular cell segmentation algorithms, including QuPath's built-in algorithm, StarDist and Cellpose. To evaluate segmentation performance, we calculated AP50, AP75 and measured 17 morphological and staining-related features for all detected cells. We compared these measurements to the gold standard of manual segmentation using the Kolmogorov-Smirnov test. 
    
    \textbf{Results:}
    Cyto R-CNN achieved an AP50 of 58.65\% and an AP75 of 11.56\% in whole-cell segmentation, outperforming all other methods (QuPath $19.46/0.91\%$; StarDist $45.33/2.32\%$; Cellpose $31.85/5.61\%$).  
    Cell features derived from Cyto R-CNN showed the best agreement to the gold standard ($\overline{D} = 0.15$) outperforming QuPath ($\overline{D} = 0.22$), StarDist ($\overline{D} = 0.25$) and Cellpose ($\overline{D} = 0.23$). 
    
    \textbf{Conclusion:}
    Our newly proposed Cyto R-CNN architecture outperforms current algorithms in whole-cell segmentation while providing more reliable cell measurements than any other model. This could improve digital pathology workflows, potentially leading to improved diagnosis. Moreover, our published dataset can be used to develop further models in the future.
\end{abstract}

\vspace{0.5cm}

  \end{@twocolumnfalse} 
]


\section{Introduction}
\label{sec:introduction}

Advances in artificial intelligence (AI) and digital pathology have revolutionized medical research. Neural networks have made it possible to predict a tumor's malignancy and prognosis \citep{RN204} and can even detect genetic differences just from histological images \citep{RN67}. But understanding the reasoning of these neural networks is not always straighforward. This is why explainability is such an important factor, both for research purposes as well as for bringing AI into clinical routine \citep{RN229}. As shown by \citet{RN230} automated cell segmentation can help fill that gap in explainability, because the cell nucleus and cytoplasm contains important morphological information. If this information can be automatically extracted from histological images via cell segmentation, this can potentially offer explanations for the predictions of a neural network.  

A lot of advancements have already been made in this field. Readily available tools such as QuPath \citep{RN9}, StarDist \citep{RN125} and Cellpose \citep{RN135} are commonly used to perform automated nucleus segmentation. However, there is a lack of methods that can reliably assist in cytoplasm segmentation in bright-field images. Therefore, many researchers still resort to the heuristic method of expanding the nucleus by a fixed number of pixels to obtain an object mask for the whole cell. This "cell expansion" \citep{RN232} is implemented for example in the popular software tool QuPath and has been used in many clinical research papers: \citet{RN253} used it in 2022 to analyze macrophages in Covid-19. \citet{RN213} used it to determine the presence of biomarkers in cells. And \citet{RN211} used it in 2023 to study a tumor's microenvironment.

Despite its popularity, cell expansion generally does not perform well on bright-field histological images as will be shown in this paper. Morphological measurements become inaccurate and staining information obtained from these segmentations is not reliable either. As a result, any statistical model built on these values will also be unreliable. 

For this reason, we developed a cell segmentation algorithm that can segmentate both the nucleus and the cytoplasm in bright-field histological images. We conducted a number of experiments to compare our method against QuPath, StarDist, CellPose and the gold standard of manual segmentation.

\section{Material and Methods}
\label{section:MaterialAndMethods}

We developed a new architecture based on Mask R-CNN for whole-cell segmentation in hematoxylin-eosin (HE) stained bright-field histological images. We then compared this architecture against cell segmentation algorithms from StarDist, Cellpose and QuPath.
To enable a fair comparison, all methods needed to be evaluated on the same dataset. To the authors' knowledge, there is no publicly available dataset that contains both nucleus and cytoplasm annotations in HE images. For this reason, we created a new dataset which is being published alongside this paper.
Using this new dataset, the models have been evaluated under different categories. First, the segmentation accuracy has been compared using the standard measures AP50 and AP75. In a second step, the predicted cell segmentations have been imported into QuPath, where a number of cell features were measured. These cell features were then compared against the gold standard measurements.

\subsection{Dataset}
\label{section:Dataset}

We created a dataset of nuclei and cell annotations using publicly available images of head and neck squamous cell carcinoma (HNSCC) from the CPTAC dataset \citep{RN199, RN194}. Nuclei and cells have been manually annotated in QuPath (version 0.4.3) by J.R. and K.X. While nuclei were generally easy to annotate, the cell membrane was not always visible due to the nature of the stain. In such cases, only the cell's nucleus was annotated. All annotations were reviewed by a third investigator B.P. and finally reviewed and approved by a senior pathologist T.B. The resulting annotations were then exported from QuPath as COCO-compatible JSON files. The  images were exported in patches of 256x256px at a resolution of 0.5$\mu m$/px. 

As the CPTAC dataset contains whole slide images from multiple facilities, the staining intensities vary between images. To eliminate these differences, all images were normalized using the Macenko algorithm \citep{RN89}. This is a standard practice used to facilitate training of deep learning models \citep{RN217}. The resulting images were then split into a training (70\%), validation (15\%) and test set (15\%) To avoid overfitting, we made sure not to allocate images from the same patient to different subsets.

\begin{figure}[h]
    \centering
    \includegraphics[width=0.23\textwidth]{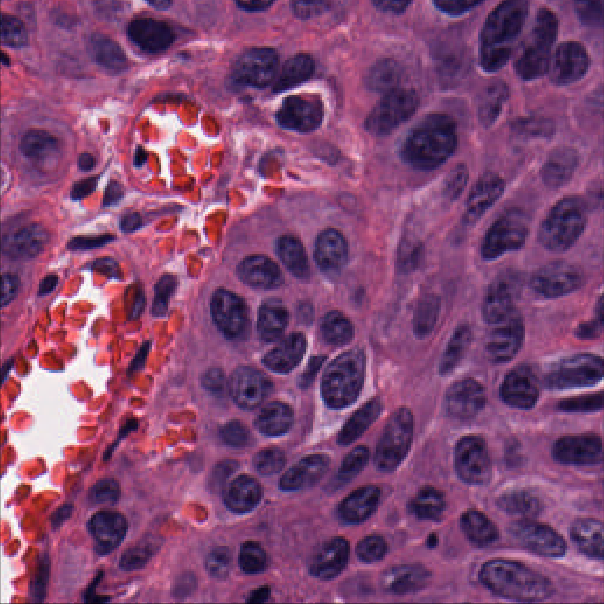}
    \includegraphics[width=0.23\textwidth]{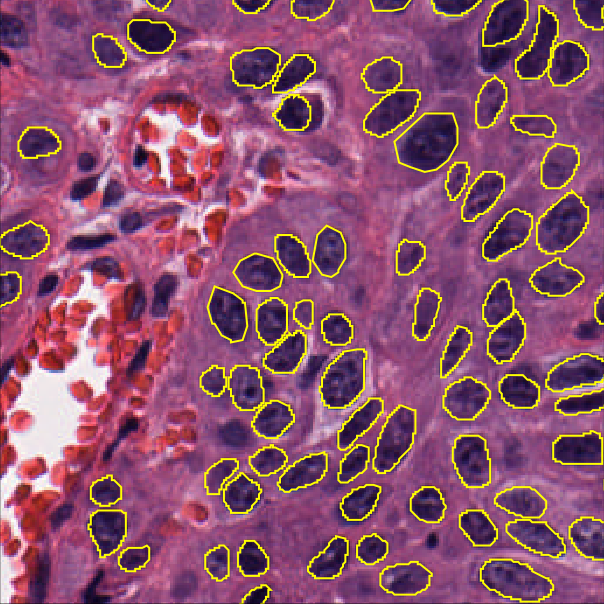}
    \includegraphics[width=0.23\textwidth]{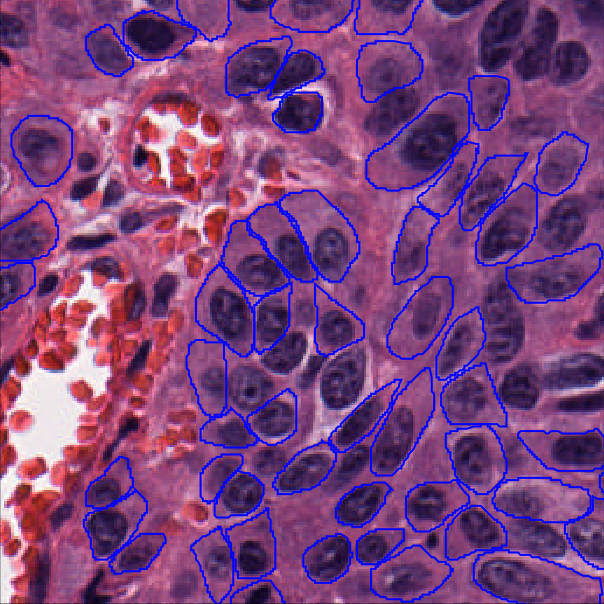}
    \includegraphics[width=0.23\textwidth]{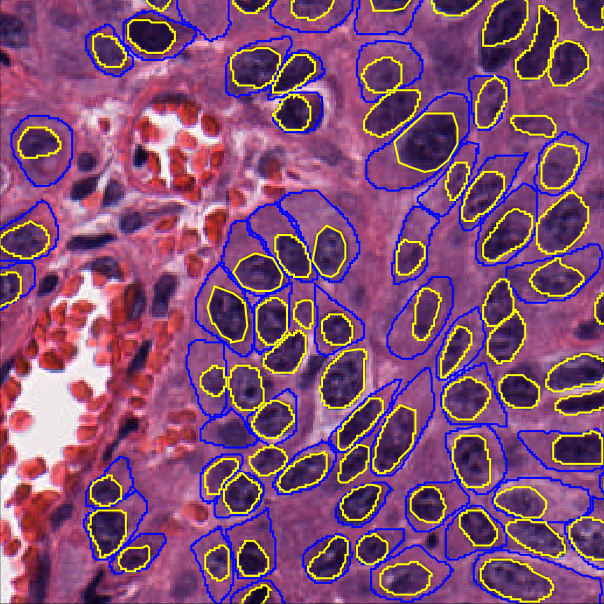}
\caption{A sample image from the CytoNuke dataset. Tumor nuclei annotations are shown in yellow, tumor cell annotations are shown in blue. Not every nucleus annotation has a corresponding cell annotation, since cell boundaries are not always clearly distinguishable.}
\label{fig:dataset}
\end{figure}

The resulting dataset contains 3,991 tumor nuclei and 2,607 tumor cell annotations. All annotations were performed manually in a quality-controlled setting. Figure \ref{fig:dataset} shows an example image from our dataset with corresponding annotations.

\subsection{Models}
\subsubsection{QuPath}
QuPath is a popular software tool to analyze whole slide images and is widely used in pathological research \citep{RN9, RN252}. It provides a built-in cell and nuclei segmentation algorithm, which we used as a baseline for our experiments.

The nucleus detection within QuPath is built upon the watershed algorithm, which utilizes the fact that a nucleus appears darker than its surrounding cytoplasm when stained by HE. Using this information, the algorithm considers adjacent dark pixels as belonging to one nucleus. To detect the whole cell including its cytoplasm, QuPath implements a method known as "nucleus expansion" \citep{RN231} or "cell expansion" \citep{RN232}. This algorithm will expand each nucleus by a fixed number of pixels as long as the expansion does not intersect with the expansion of an adjacent nucleus. 

Since the watershed algorithm is sensitive to its hyperparameters, QuPath has been evaluated in two different modes: Once with default settings and once with a set of improved parameters.

The default settings in QuPath version 0.4.3 were as follows: A background radius of $8\mu m$, a sigma of $1.5 \mu m$, a minimum area of $10 \mu m^2$, a maximum area of $400 \mu m^2$, an intensity threshold of $0.1$ and a cell expansion radius of $5 \mu m$.

We heuristically tuned the parameters on our validation dataset and found the following values to deliver good results for nucleus segmentation on our dataset: A sigma of $2.5\mu m$, a minimum area of $20 \mu m^2$, a maximum area of $400 \mu m^2$ and an intensity threshold of $0.15$. Using these parameters, cell expansion has been performed with all possible expansion radii in a range from $0.5\mu m$ to $10\mu m$ and optimized on the validation dataset. The best expansion radius has been identified to be $5 \mu m$. 

\begin{figure}[t]
    \centering
    \includegraphics[width=0.46\textwidth]{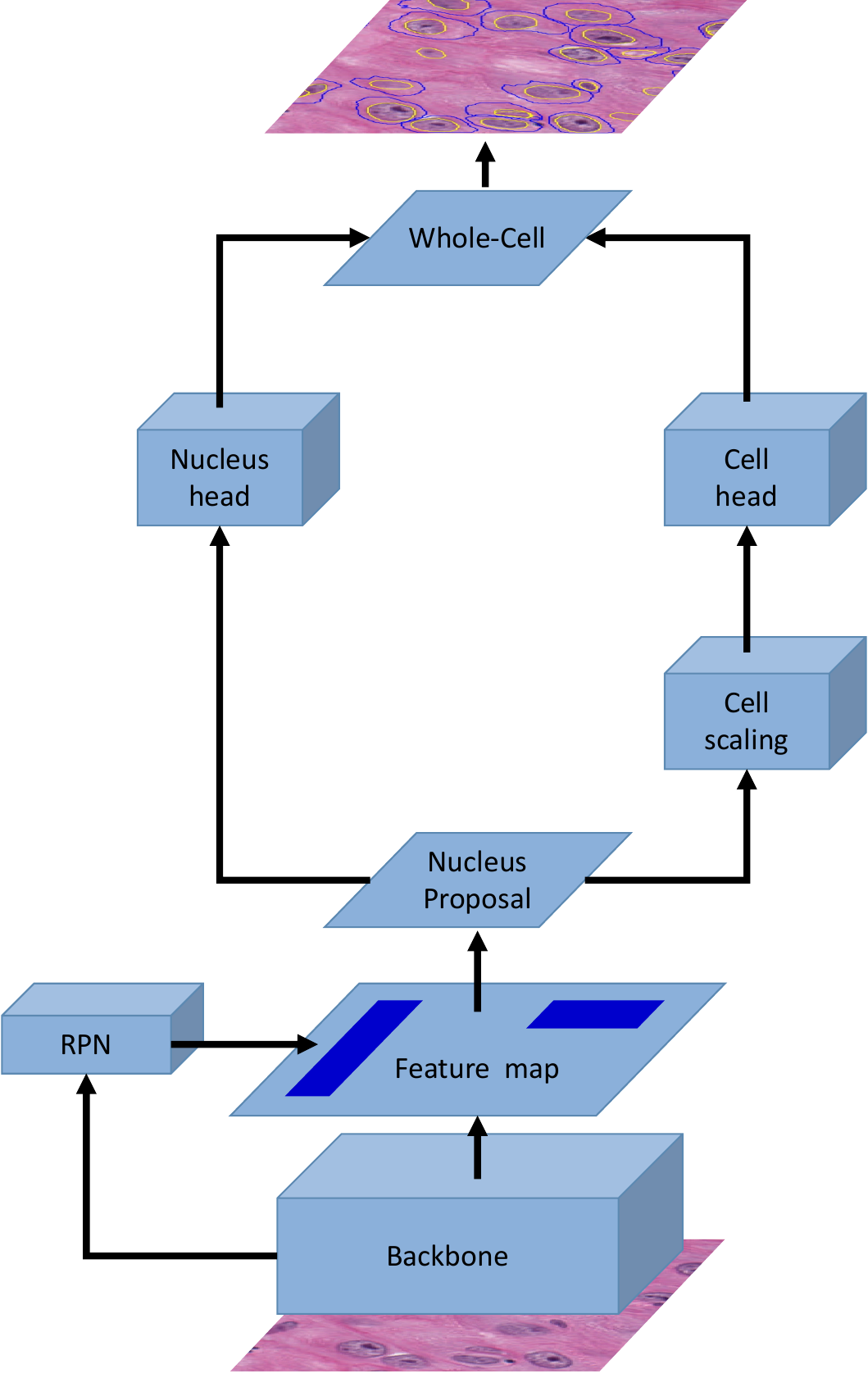}
\caption{Architectural overview of Cyto R-CNN. The backbone and RPN are trained on nuclei only. The nuclei proposals are then forwarded to two different branches. The first branch will perform a regular bounding box and mask regression for the nucleus. The second branch will scale the nucleus proposal and perform mask regression for the whole cell, including the cytoplasm. Both branches are then combined to generate instance segmentations for cell and nucleus at the same time.}
\label{fig:cyto_rcnn_architecture}
\end{figure}

\subsubsection{StarDist}

StarDist is a U-Net based CNN model developed specifically for the purpose of nucleus detection \citep{RN125}. It is built on the assumption that a nucleus' shape can be approximated by a star-convex polygon. For each pixel, StarDist predicts two values: First, the probability that the given pixel belongs to a nucleus. Second, the estimated distance towards the boundary of the nucleus for a total of 32 different directions.

For our experiments, we used the original tensorflow implementation \citep{RN125} and finetuned it with our dataset. To avoid overfitting during training, the input data has been augmented with random flips and rotations. 
We also experimented with both training the network from scratch as well as using the pre-trained model \textit{2d\_versatile\_he} which the authors published for segmentation tasks in HE images. Using the HE-pre-trained model proved beneficial in our experiments.

After training, the different models were evaluated on the validation dataset. The best model was then combined with traditional nucleus expansion. For the nucleus expansion, we again experimented with different expansion radii from $0.5\mu m$ to $10\mu m$. Evaluation on the validation dataset identified the best expansion radius to be $5.5 \mu m$.

\subsubsection{Cellpose}

Cellpose is a second U-Net based segmentation model. Unlike StarDist however, Cellpose was originally designed to work with immunohistological images, where nucleus and cytoplasm were stained in different bright colors \citep{RN135}. This special coloring allows Cellpose to perform color gradient tracking to derive the object masks.
Since its invention, Cellpose has shown potential to generalize beyond its initial scope of immunofluorescence images \citep{RN236}. For this reason, we decided to include it in our experiments as well.

We experiment with training it from scratch as well as finetuning it starting from a number of pre-trained models. The input data was again augmented with random flips and rotations. We also experimented with all possible combinations of color channels that serve as a hyperparameter for the Cellpose model.

Out of all the pre-trained Cellpose models, the model \textit{CPx} performed best, both in nucleus and whole-cell segmentation measured by AP50 and AP75. It delivered optimal performance when setting the nucleus channel to $3$ and the cytoplasm channel to $0$. $CPx$ has then been finetuned on our dataset.

\subsubsection{Cyto R-CNN}
\label{section:cytorcnn}

We propose a new architecture based on Mask R-CNN that allows us to segment both the nucleus as well as the whole cell. An illustration of its design is shown in figure \ref{fig:cyto_rcnn_architecture}.

Mask R-CNN is a general-purpose instance segmentation model \citep{RN49} which has also been applied to nucleus segmentation tasks in the past. It works as a two-stage system. The first stage extracts a number of rectangular regions of interest (ROIs) from the image, that are likely to contain an object. The second stage is then responsible for classifying the object and predicting its binary object mask. This approach has proven successful in many general-purpose instance segmentation tasks, including nucleus segmentation \citep{RN237, RN239}.

Just like a regular Mask R-CNN the first stage of Cyto R-CNN is a regional proposal network (RPN) to predict rectangular ROIs. However, the RPN is only trained on nucleus annotations. Instead of a single second stage, Cyto R-CNN then proceeds in two branches: One for the nucleus and one for the whole cell. The nucleus branch consists of the regular Mask R-CNN steps: First a bounding box regression, then a binary mask prediction. The cell branch on the other hand has to first create a regional proposal for the cytoplasm. This is achieved by scaling the nucleus ROI by a fixed percentage. This is a promising heuristic, since the nucleus is always strictly surrounded by cytoplasm. The resulting cell-ROI is then passed forward to a regular mask head.

\begin{figure*}[bt]
    \centering
    \includegraphics[width=\textwidth, height=0.4\textheight]{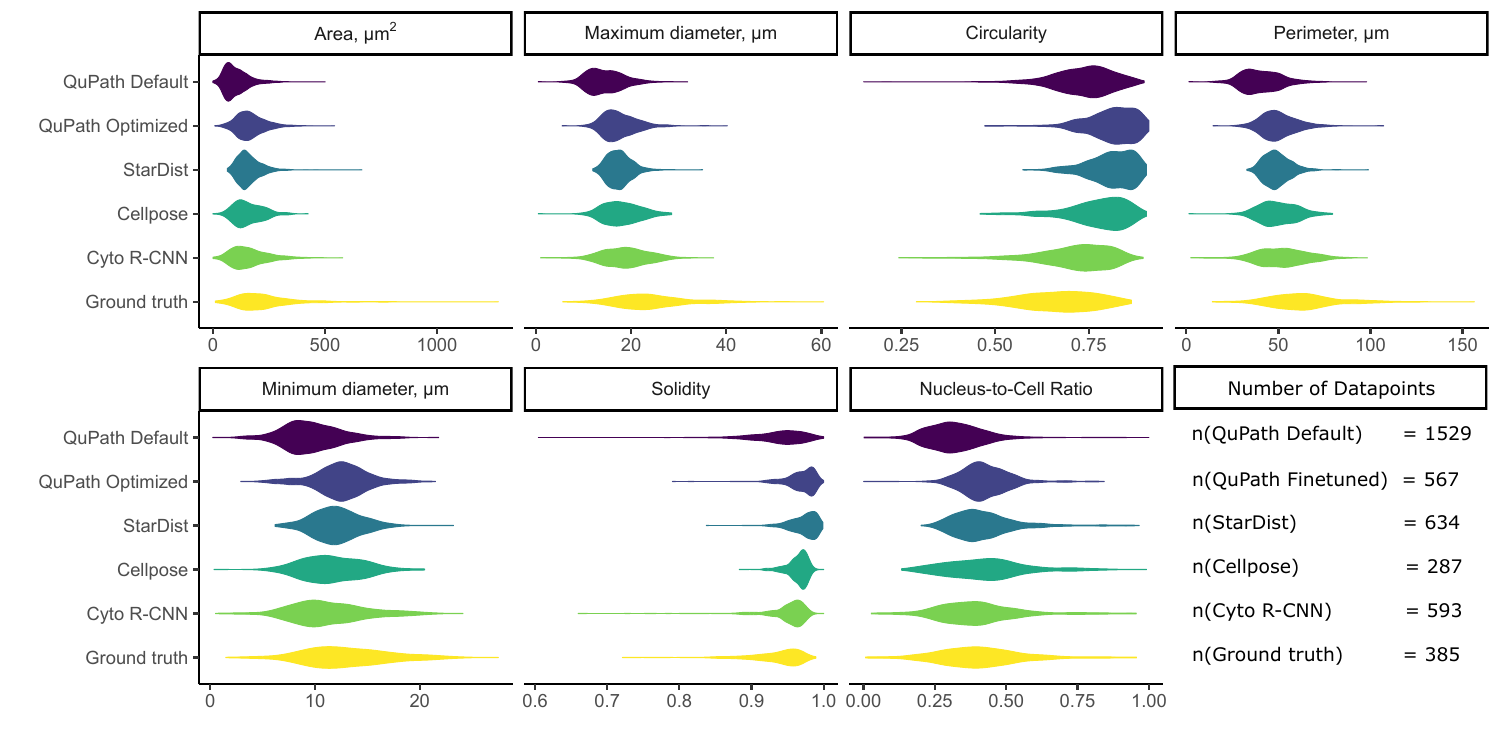}
\caption{Measurements of morphological whole-cell features as obtained via different segmentation methods. The measurements have been obtained by first converting segmentation masks into GeoJSON files, importing them into QuPath and then using built-in functionalities to calculate shape and staining features.}
\label{fig:violin_plot_morphology}
\end{figure*}

\begin{table*}[htbp]
\caption{\label{table:kolmogorov_smirnov}Test statistics $D$ of the Kolmogorov-Smirnov test between each model and the gold standard. } 

\begin{tblr}{
    colspec = { *{2}{c} *{12}{X[c]} },
    hlines, 
    vlines,
    process=\grhighlowColor
}

Feature & QuPath Default & QuPath Finetuned & StarDist & Cellpose & Cyto R-CNN \\
Area & 0.54 & 0.26  & 0.32  & 0.27 & 0.25  \\
Perimeter & 0.60 & 0.42 & 0.49 & 0.40 & 0.29 \\ 
Circularity & 0.31 & 0.61 & 0.61 & 0.47 & 0.24 \\
Solidity & 0.11 & 0.48 & 0.55 & 0.46 & 0.24 \\ 
Max. diameter & 0.64 & 0.48 & 0.56 & 0.42 & 0.34 \\
Min. diameter & 0.39 & 0.17 & 0.21 & 0.19 & 0.20 \\ 
Nucleus-to-Cell Ratio & 0.17 & 0.37 & 0.41 & 0.51 & 0.17 \\ 
Hematoxylin Median & 0.11 & 0.24 & 0.13 & 0.19 & 0.14 \\
Hematoxylin Mean & 0.10 & 0.17 & 0.13 & 0.19 & 0.11 \\
Hematoxylin Std. Dev. & 0.05 & 0.17 & 0.12 & 0.05 & 0.07 \\
Hematoxylin Max. & 0.17 & 0.09 & 0.05 & 0.07 & 0.03 \\
Hematoxylin Min. & 0.13 & 0.13 & 0.10 & 0.19 & 0.04 \\
Eosin Median & 0.10 & 0.16 & 0.15 & 0.19 & 0.13 \\
Eosin Mean & 0.07 & 0.17 & 0.15 & 0.14 & 0.11 \\
Eosin Std. Dev. & 0.06 & 0.17 & 0.12 & 0.10 & 0.08 \\
Eosin Max. & 0.07 & 0.06 & 0.06 & 0.07 & 0.07 \\
Eosin Min. & 0.11 & 0.11 & 0.08 & 0.08 & 0.05 \\
\end{tblr}
\end{table*}

\begin{figure*}[bt]
    \centering
    \includegraphics[width=\textwidth, height=0.4\textheight]{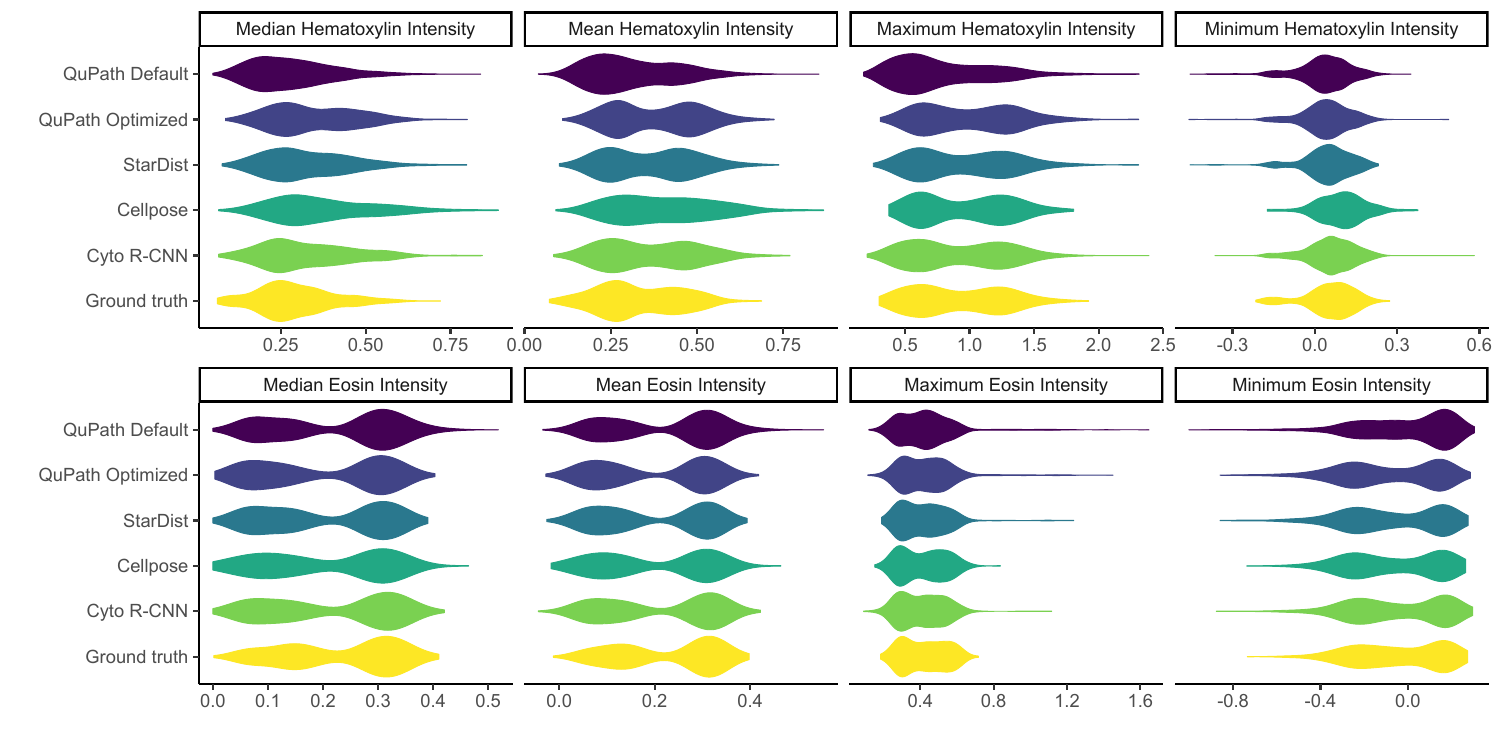}
\caption{Measurements of whole-cell staining features resulting from different segmentation methods. The above measurements have been obtained by first converting segmentation masks into GeoJSON files, importing them into QuPath and then using built-in functionalities to calculate shape and staining features.}
\label{fig:violin_plot_staining}
\end{figure*}

In a way, this design can be considered an enhancement of the original nucleus expansion in QuPath. While there is still a heuristic at play to generate the cell ROI, a neural network is then used to refine this initial guess.

As Mask R-CNN is a large network, there are a number of hyperparameters that can be tuned.
The backbone of Cyto R-CNN has been set to \textit{mask\_rcnn\_R\_101\_FPN}, pre-trained on COCO \citep{RN251} as provided by detectron.
Training data was augmented with random flips and rotations.
Different learning rates were evaluated using an exponential learning rate decay.
Different cell scaling factors were evaluated as well. The optimal cell scaling factor for our dataset was determined to be $2.0$.
For the ROI heads, the non-maximum suppression was tuned due to the relatively large number of small objects in each image. A threshold of $0.3$ delivered best results in our experiments.

\subsection{Computational setup}
Our experiments were performed on GPUs of type V100-SXM2. The algorithms were implemented using python 3.9, detectron2 \citep{RN154}, pytorch and tensorflow.

\subsection{Statistical Analysis}
All methods were evaluated in their performance using class-specific AP50 and AP75. These two endpoints are commonly used in the literature \citep{RN248, RN225}, as they allow for insights not only into the detection accuracy, but also into the segmentation quality of a model. A model with high AP50, but low AP75 will be expected to detect objects quite well, but would fail to generate segmentation masks that accurately match their shape.

The segmentation masks produced by each model were subsequently imported into QuPath to measure a set of cell features related to morphology and staining. The features we chose for our comparison were: Cell area, perimeter, circularity, solidity, minimum and maximum diameter, cell-to-nucleus ratio, HE staining (minimum, maximum, median, mean, standard deviation). This specific set of features was chosen to resemble what is commonly used in the literature \citep{RN244, RN240, RN245, RN243, RN242}. 

The resulting measurements were then compared to the ground truth measurements using the Kolmogorov-Smirnov test. We used its test statistic $D$ to quantify the similarity between the measurements derived from the different models and the measurements resulting from manual annotations. This statistical analysis was implemented in R (v4.2.2).

\section{Results}

\subsection{Segmentation accuracy}

\begin{table}[htbp]
\caption{\label{table:performance_results}Average precision on the test dataset}
    \resizebox{\columnwidth}{!}{
        \begin{tblr}{
          hlines,
          vlines,
        }
        Model                & {AP50\\ Nucleus} & {AP75\\ Nucleus} & {AP50\\ Cell} & {AP75\\ Cell} \\
        {QuPath Default}       & 22.95\%          & 6.85\%           & 11.12\%       & 0.28\%        \\
        {QuPath Finetuned}     & 35.24\%          & 11.07\%          & 19.46\%       & 0.91\%        \\
        {Cellpose}             & 48.35\%          & 23.84\%          & 31.85\%       & 5.61\%        \\
        {StarDist}             & 70.36\%          & \textbf{47.24}\%          & 45.33\%       & 2.32\%        \\
        {Cyto R-CNN}           & \textbf{78.32}\%          & 42.54\%          & \textbf{58.65}\%       & \textbf{11.56}\%       
        \end{tblr}
    }
\end{table}

Table \ref{table:performance_results} lists the AP50 and AP75 for whole-cell and nucleus segmentation. Cyto R-CNN achieves the highest accuracies for whole-cell segmentation, both in AP50 (58.65\%) and AP75 (11.56\%). For nucleus segmentation, Cyto R-CNN achieves the highest AP50 (78.32\%) and StarDist the highest AP75 (47.24\%). QuPath and Cellpose are being outperformed in all categories.

\subsection{Measurement of cell features}

The cell measurements as calculated by QuPath are visualized in figures \ref{fig:violin_plot_morphology} and \ref{fig:violin_plot_staining}. Figure \ref{fig:violin_plot_morphology} contains information about morphological features, while figure \ref{fig:violin_plot_staining} addresses the staining-related features. A tabular representation of these values can be found in supplementary table \ref{table:cell_features_median_mean_iqr}. The results of the Kolmogorov-Smirnov test can be found in \ref{table:kolmogorov_smirnov}.

The test statistic $D$ provides a quantification of how similar the distribution of predicted cell features are to the distribution of the gold standard. $D$ ranges from 0 (perfect resemblance to the gold standard) to 1 (strong deviation from the gold standard).
For the cell area, all methods are somewhat similar in their performance relative to the gold standard ($0.25 \leq D \leq 0.32$). Only the default settings in QuPath result in worse performance ($D = 0.54$).
For the cell perimeter, predictions of Cyto R-CNN ($D = 0.29$) were closer to the gold standard than all other methods. Default QuPath resulted in the worst results ($D=0.60$). 
Regarding the cell circularity, Cyto R-CNN was the most accurate method ($D=0.24$), closely followed by QuPath's default ($D = 0.31$). Other methods have been significantly worse ($0.46 \leq D \leq 0.61$).
The cell solidity was best approximated by QuPath's default ($D=0.11$). Cyto R-CNN offered the second best approximation ($D=0.24$).
For the maximum diameter, the best results were obtained by Cyto R-CNN ($D=0.34$), followed by Cellpose ($D=0.42$) and QuPath Finetuned ($D=0.48$).
The minimum diameter was best approximated by finetuned QuPath ($D=0.17$), closely followed by Cellpose ($D=0.19$), Cyto R-CNN ($D=0.20$) and StarDist ($D=0.21$). Only default QuPath was significantly worse in its performance ($D=0.39$).
For nucleus-to-cell ratio Cyto R-CNN and default QuPath were the best methods ($D=0.17$). All other methods were significantly worse ($0.37 \leq D \leq 0.51$).
Hematoxylin median and mean intensity were best predicted by default QuPath ($D_{median}=0.11$, $D_{mean}=0.10$). However, all models offered similarly good performance ($0.10 \leq D \leq 0.24$).
The cells' standard deviation of hematoxylin intensity was best approximated by default QuPath and Cellpose (both $D=0.05$) with Cyto R-CNN following closely ($D=0.07$).  
Hematoxylin minimum and maximum intensities were best approximated by Cyto R-CNN ($D=0.04$ and $D=0.03$). QuPath default, while performing good in mean intensity, results in larger deviations here ($D=0.13$ and $D=0.17$). 
The best results for eosin median and mean intensity were delivered by QuPath Default ($D_{median}=0.10$, $D_{mean}=0.07$) and Cyto R-CNN ($D_{median}=0.13$, $D_{mean}=0.11$). 
The same applies to the standard deviation of eosin intensity ($D_{qupath}=0.06$ and $D_{cytorcnn}=0.08$).
Eosin minimum and maximum intensities were best approximated by Cyto R-CNN ($D=0.06$ and $D=0.05$), but all models were generally close in performance ($0.05 \leq D \leq 0.11$).

Summarizing the results above, Cyto R-CNN's predictions show the highest similarity to the gold standard in 9 out of the 17 examined cell features. In 6 of the remaining 8 categories, Cyto R-CNN offers the second best approximation. In two categories, Cyto R-CNN provided third best results. Based on all 17 features Cyto R-CNN had the best average agreement with the ground truth ($\overline{D} = 0.15$) outperforming QuPath ($\overline{D} = 0.22$), StarDist ($\overline{D} = 0.25$) and Cellpose ($\overline{D} = 0.23$).

\section{Discussion}
\label{section:Discussion}

The proposed architecture Cyto R-CNN enables segmentation of nucleus and cytoplasm in bright-field histological images. It outperforms QuPath, StarDist and Cellpose in the whole-cell AP50 and AP75. Moreover, Cyto R-CNN's predictions are more reliable for measuring cell features than any other model.

\begin{figure*}[bt!]
    \centering
    \includegraphics[width=3.5cm]{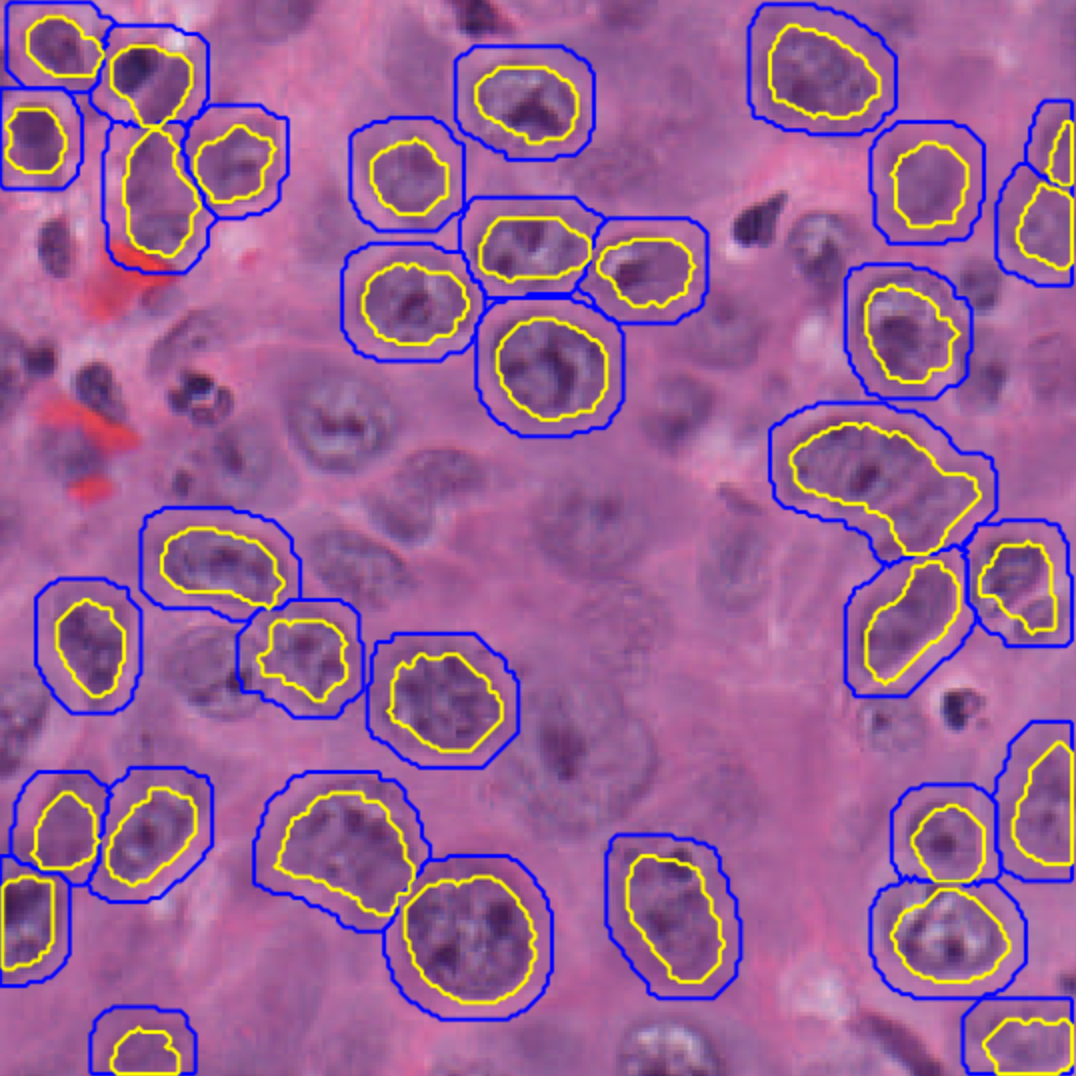}
    \includegraphics[width=3.5cm]{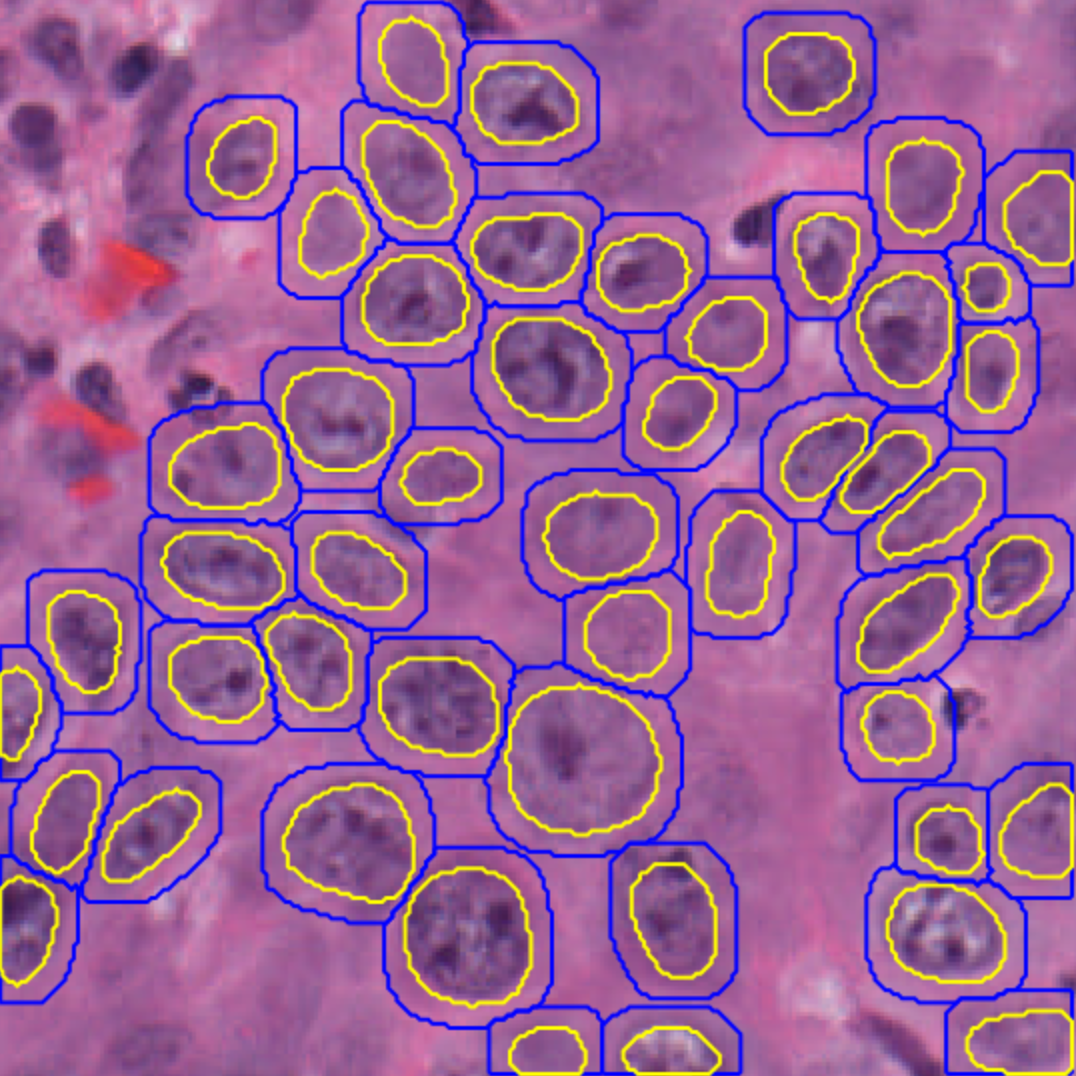}
    \includegraphics[width=3.5cm]{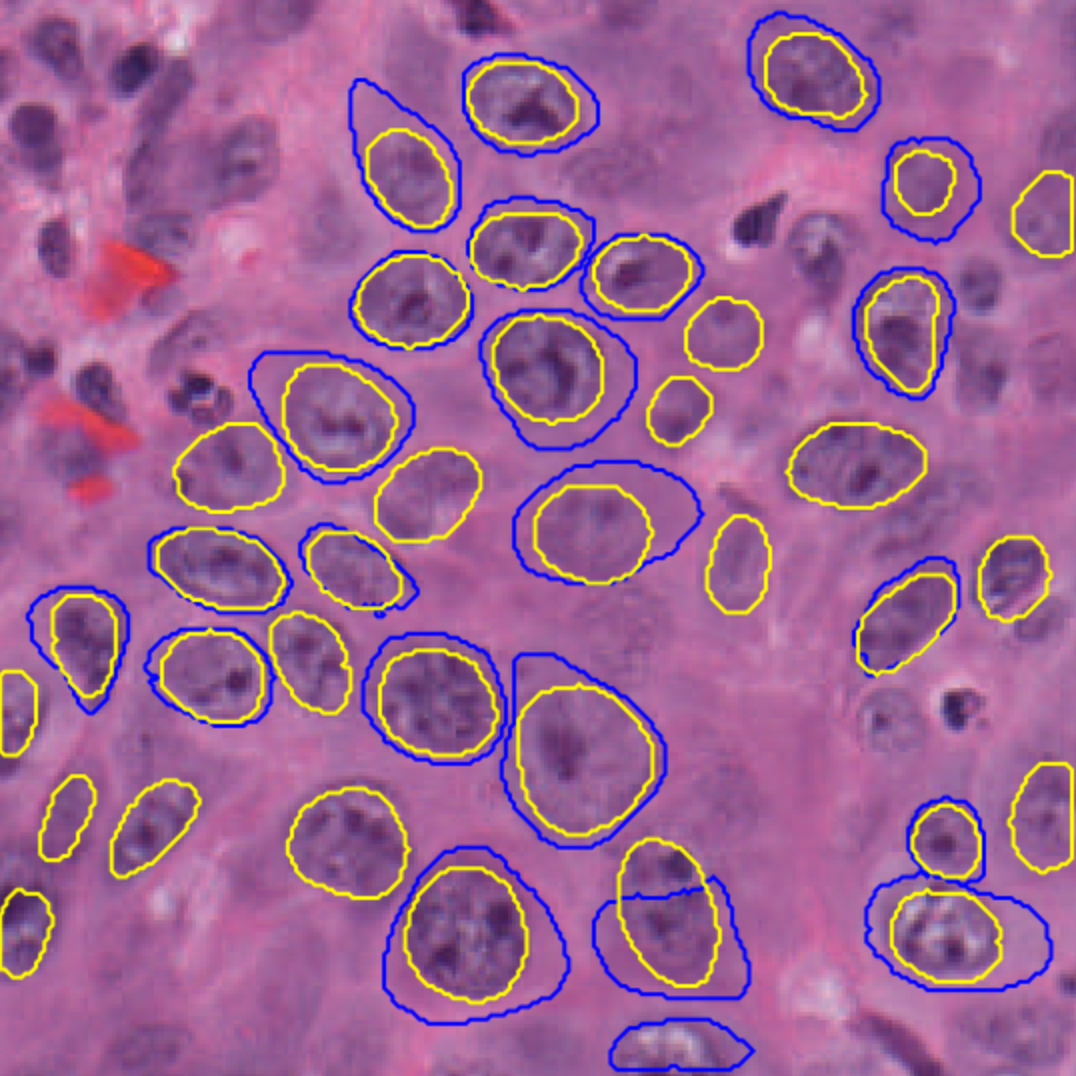}
    \includegraphics[width=3.5cm]{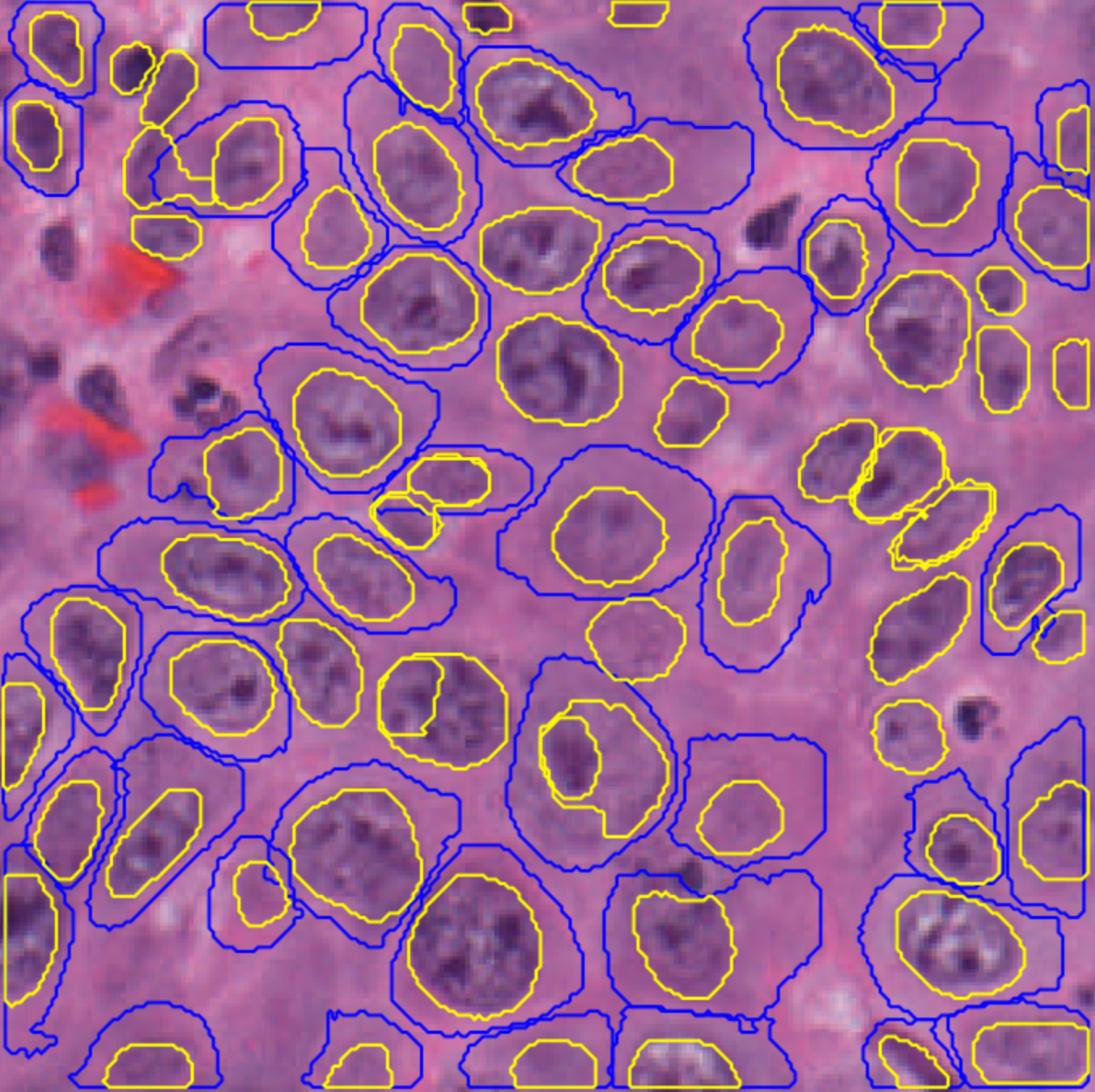}
    \includegraphics[width=3.5cm]{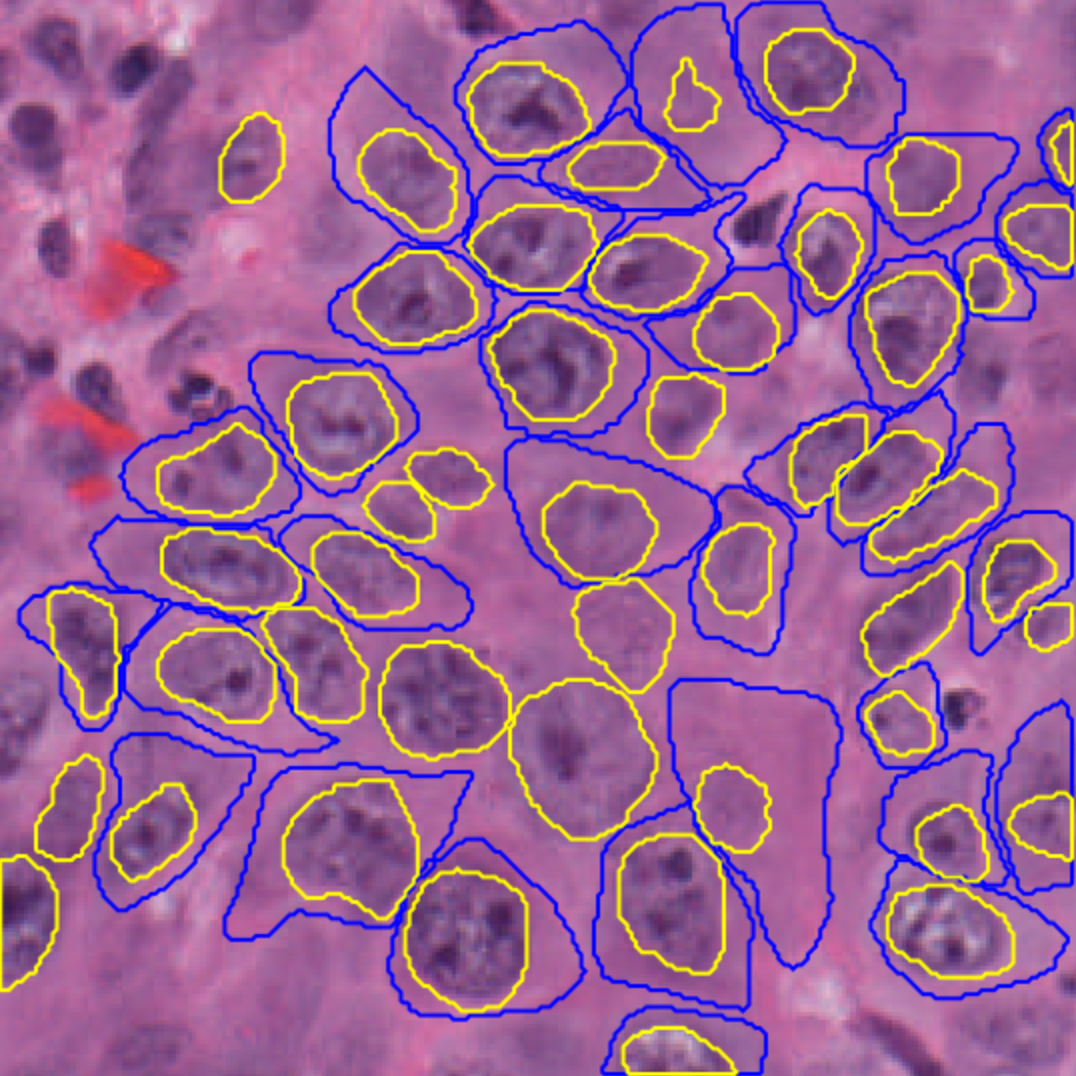}
    \includegraphics[width=3.5cm]{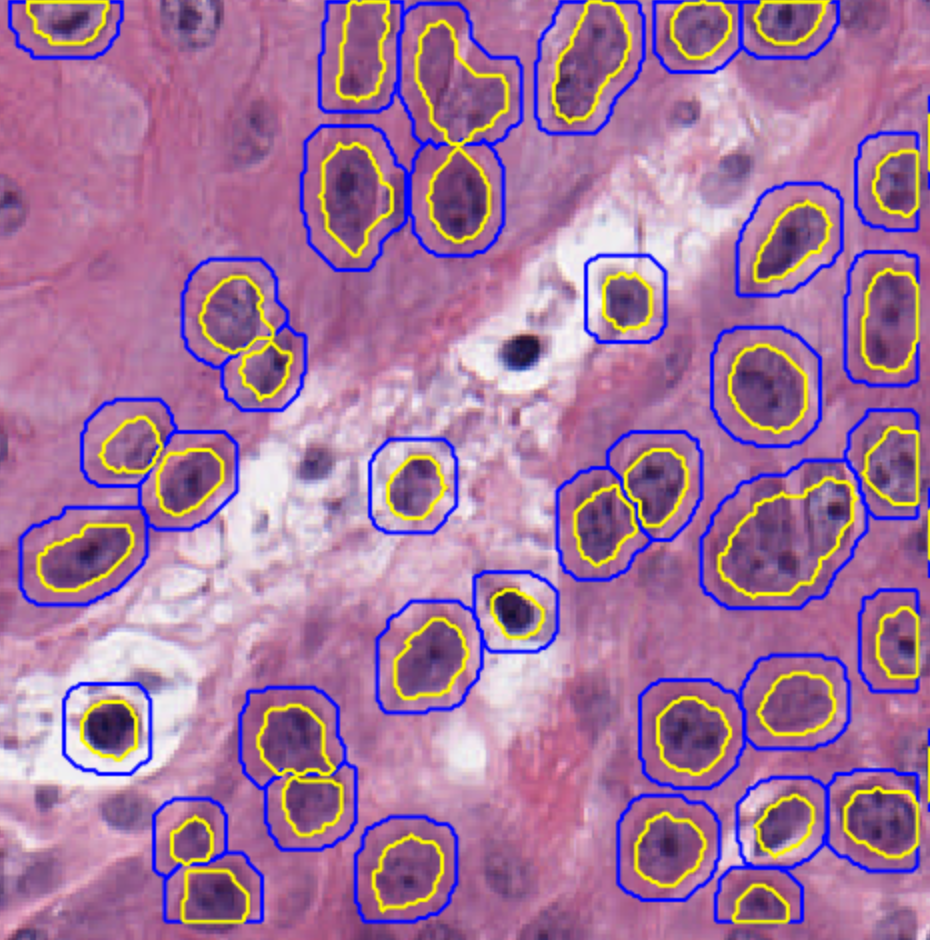}
    \includegraphics[width=3.5cm]{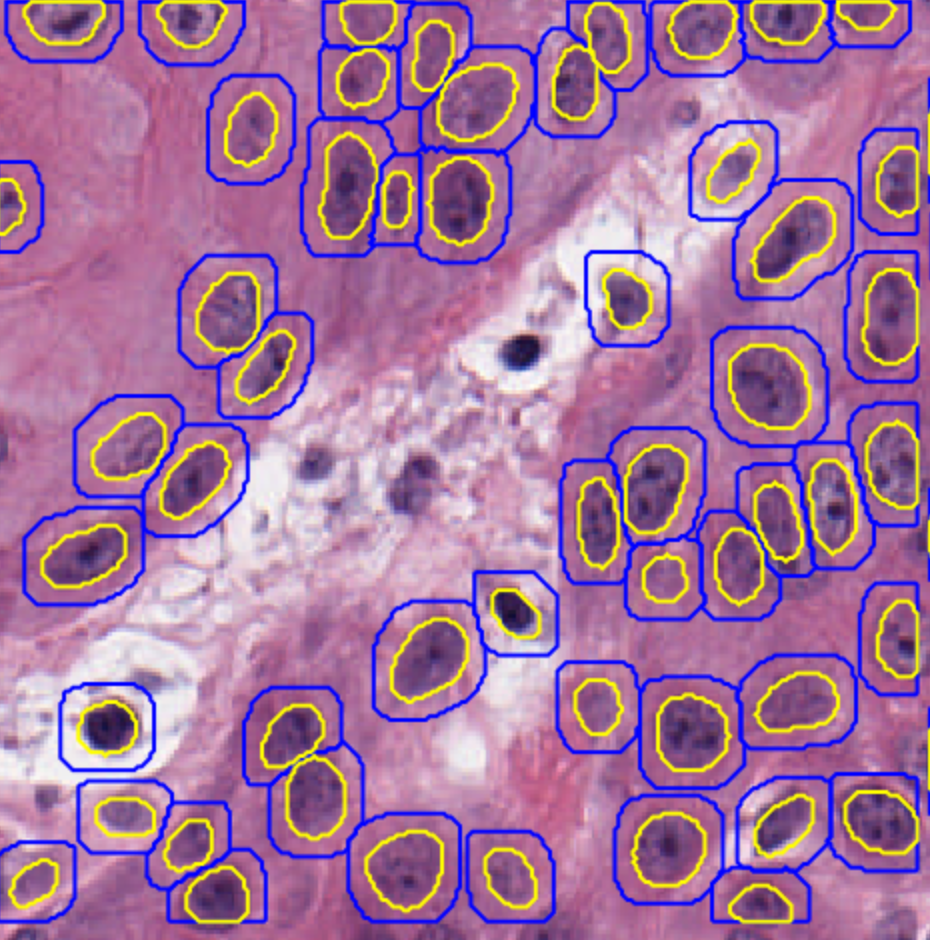}
    \includegraphics[width=3.5cm]{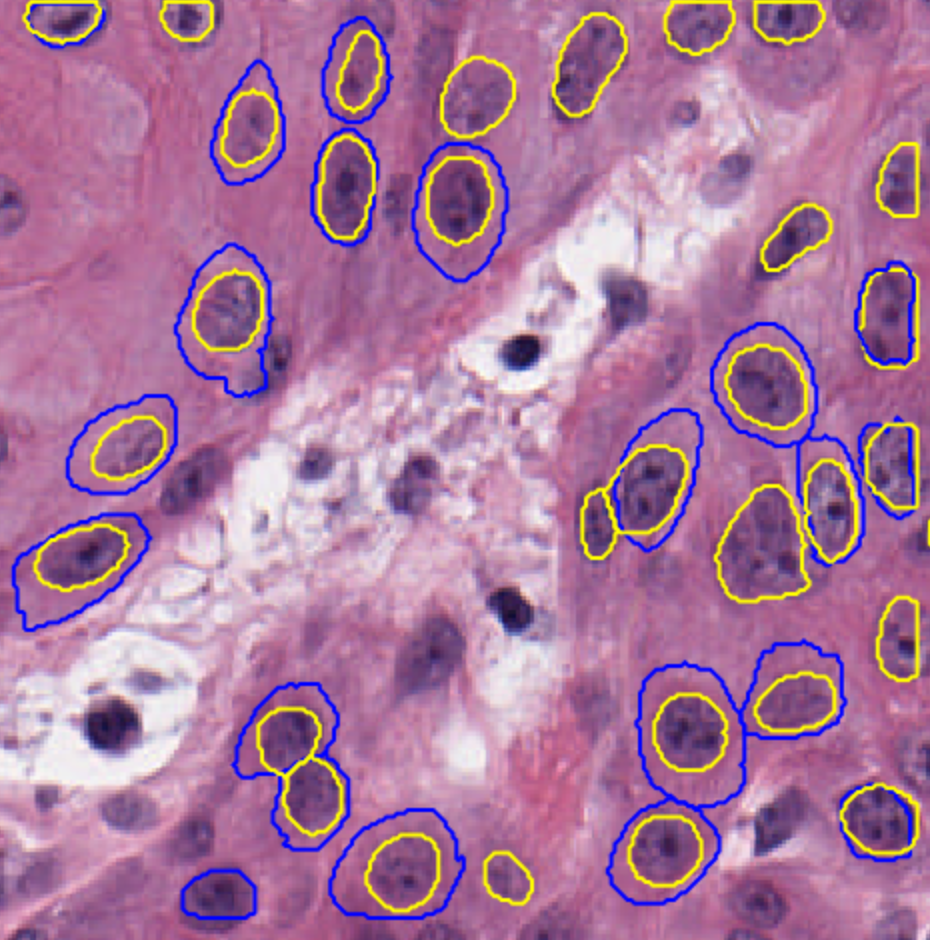}
    \includegraphics[width=3.5cm]{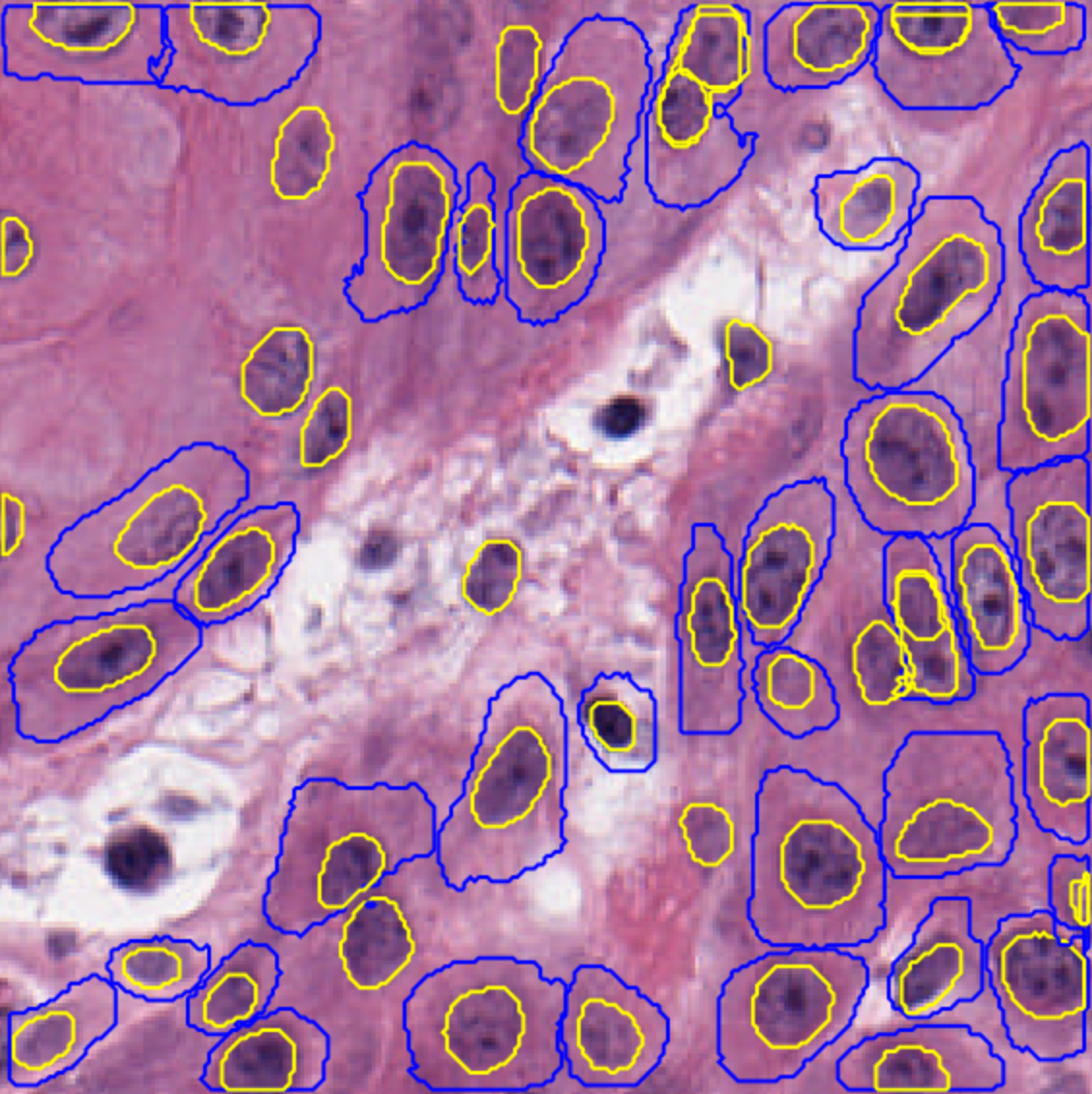}
    \includegraphics[width=3.5cm]{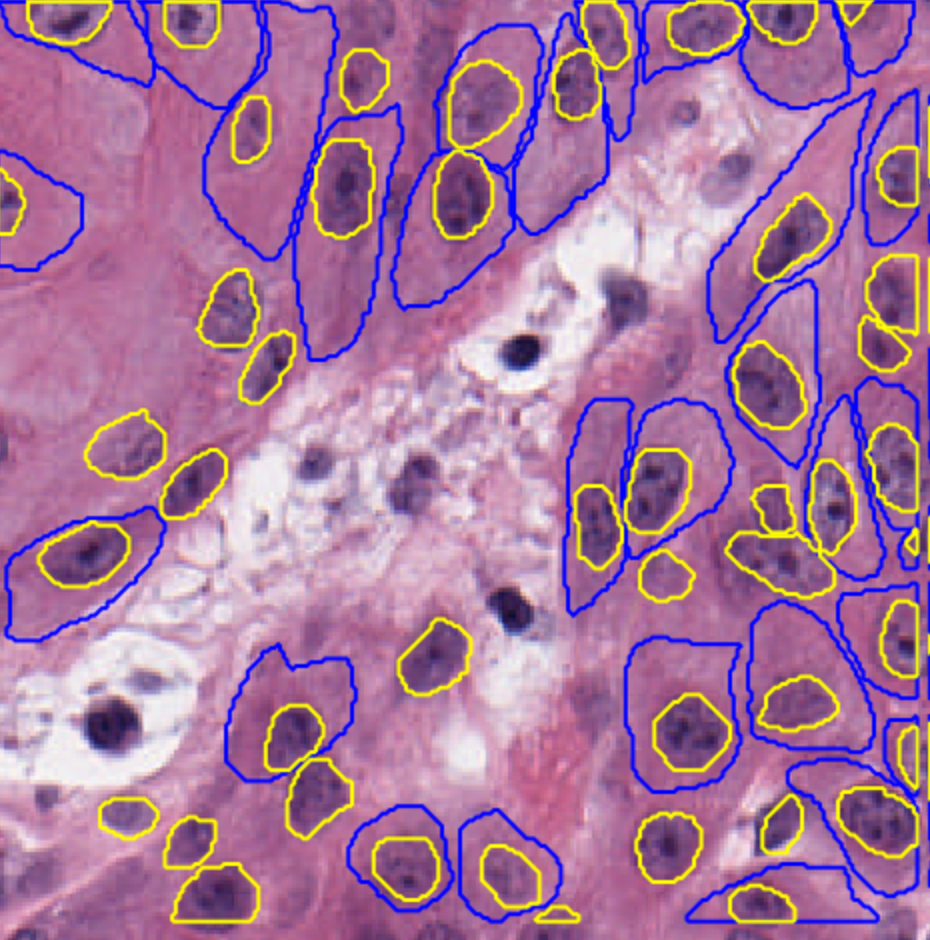}
    \includegraphics[width=3.5cm]{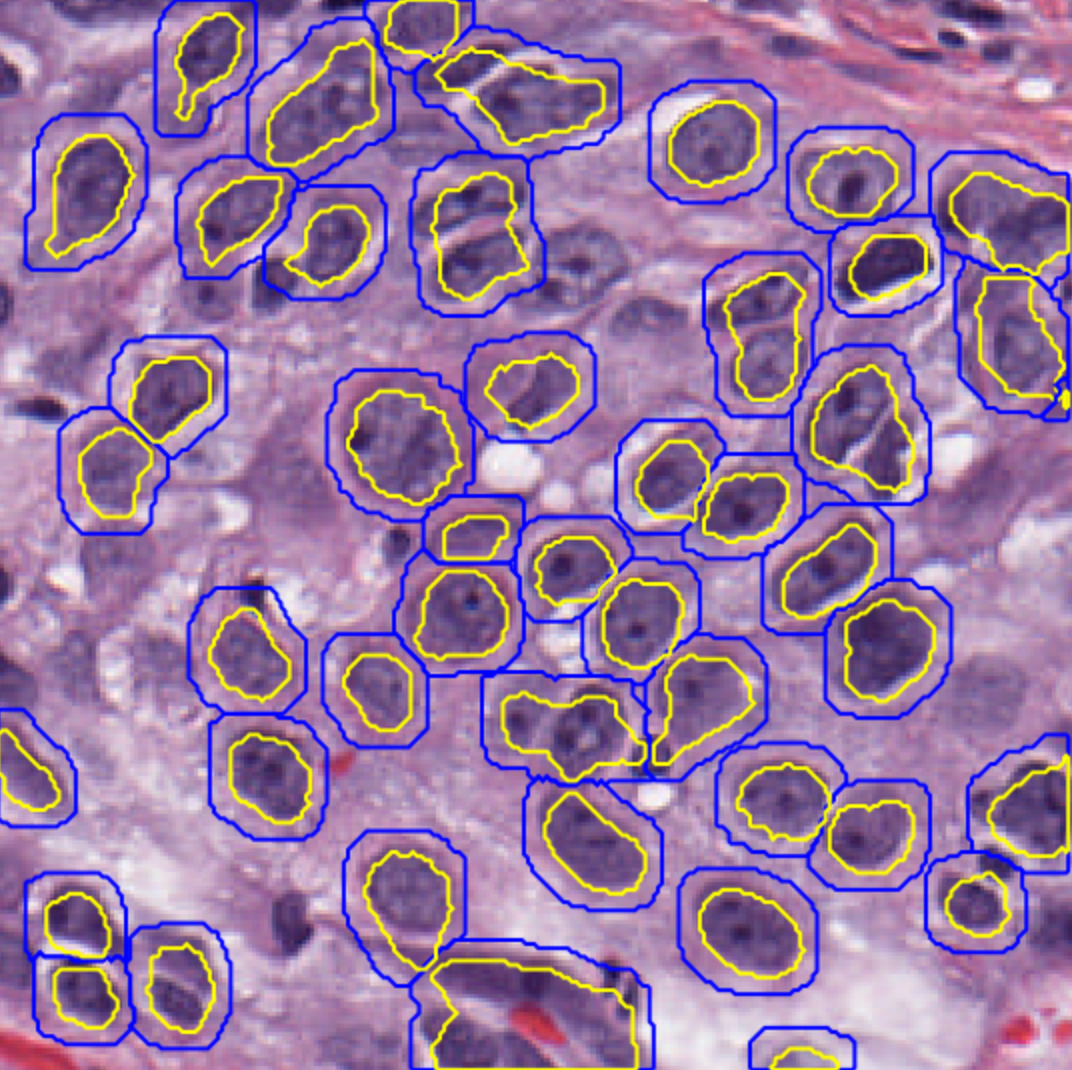}
    \includegraphics[width=3.5cm]{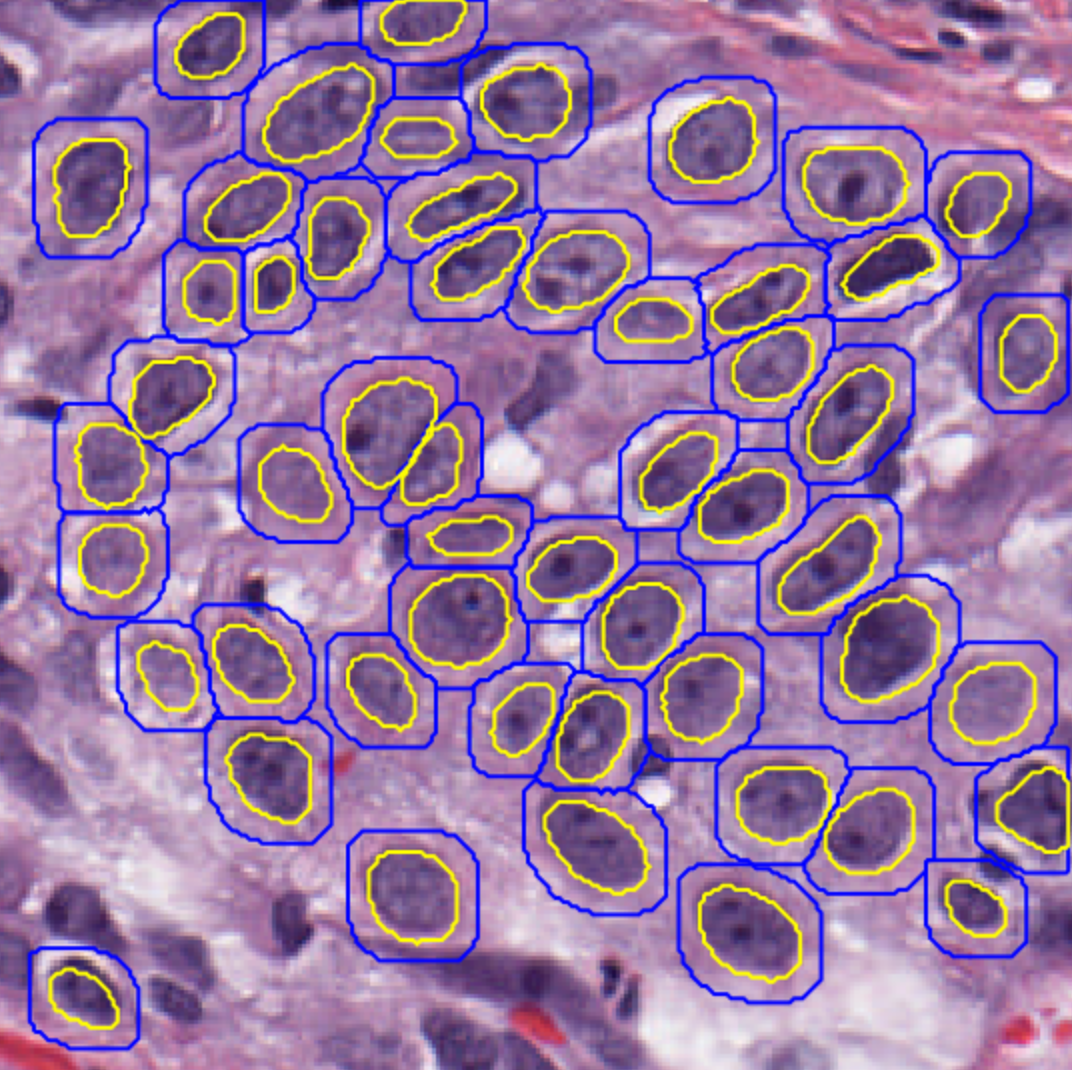}
    \includegraphics[width=3.5cm]{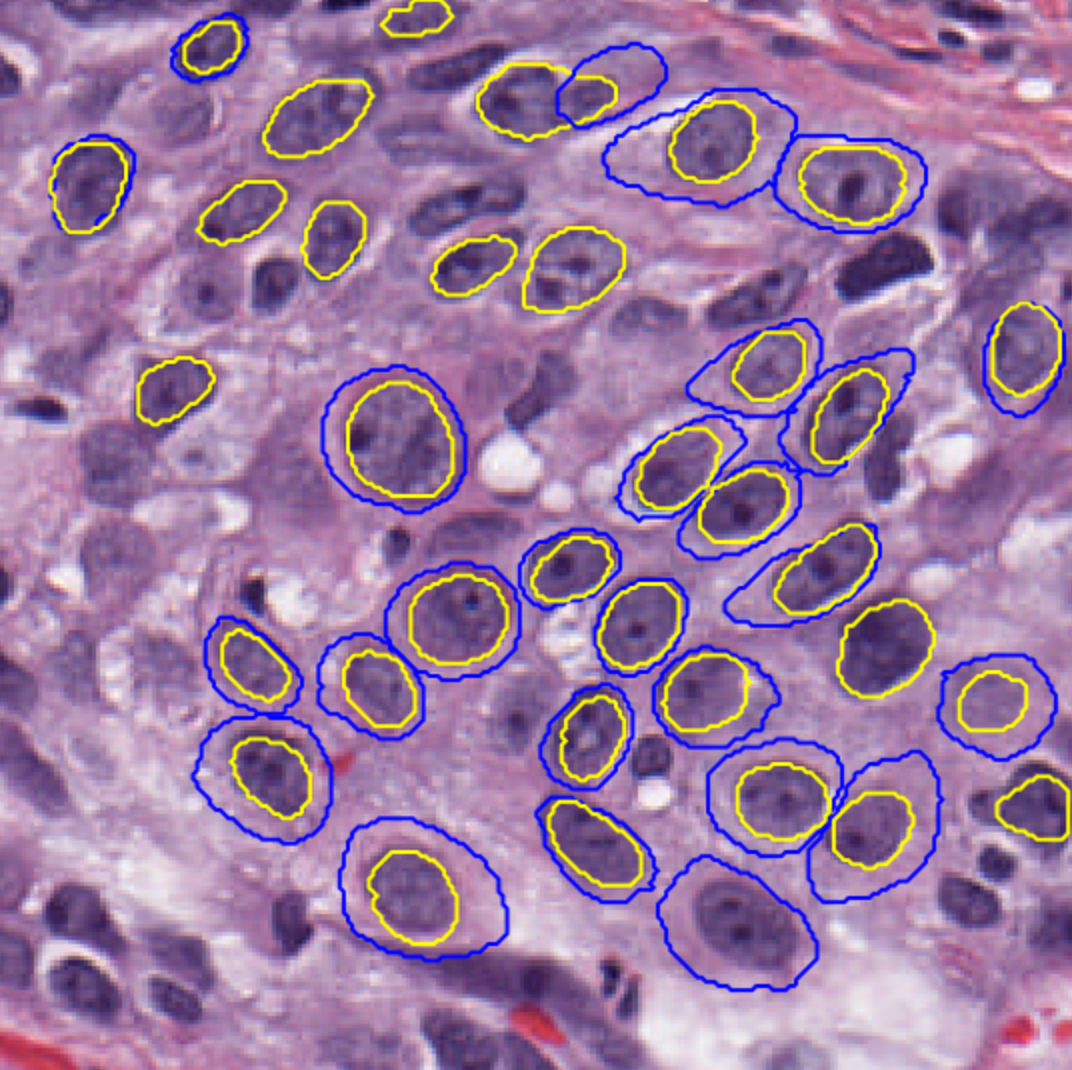}
    \includegraphics[width=3.5cm]{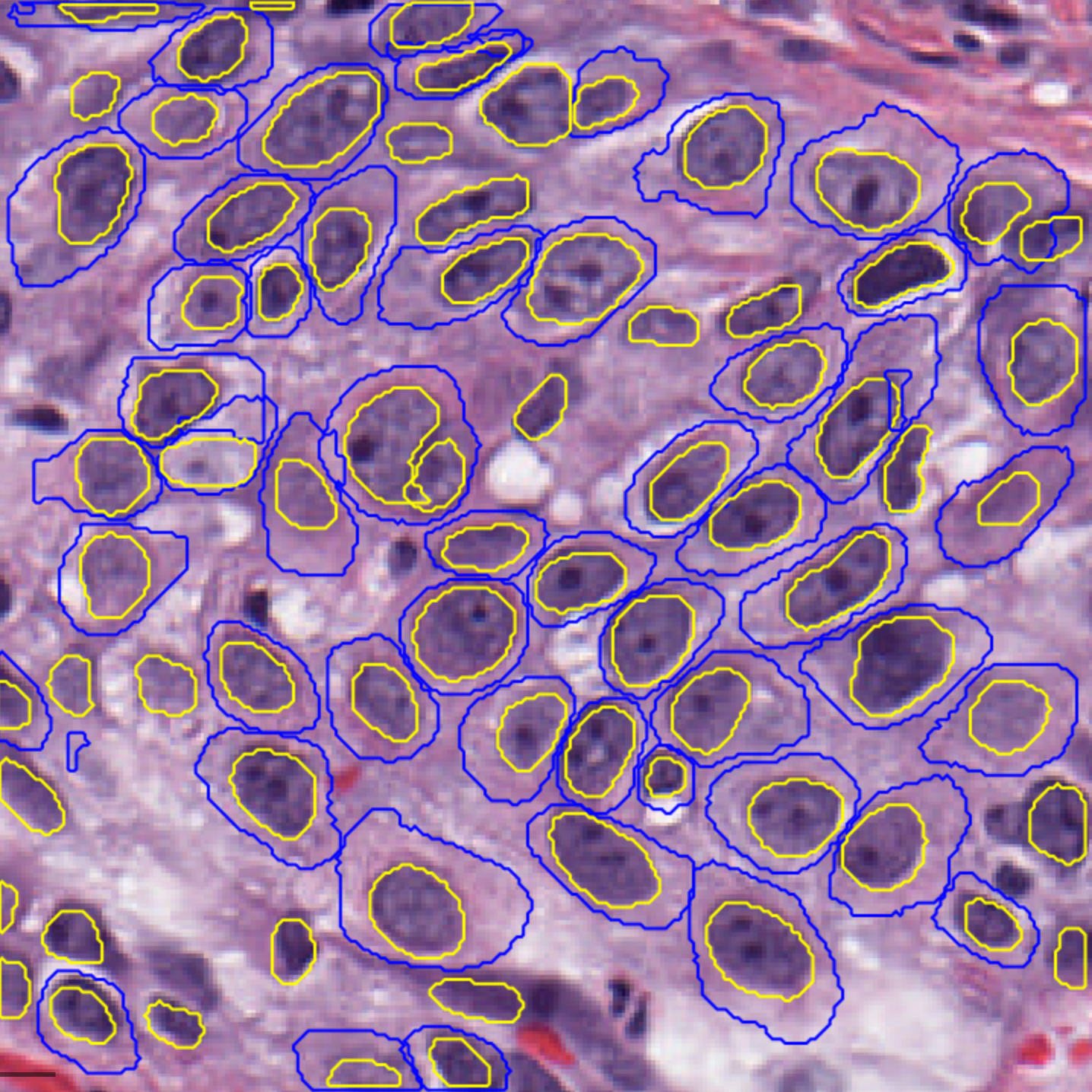}
    \includegraphics[width=3.5cm]{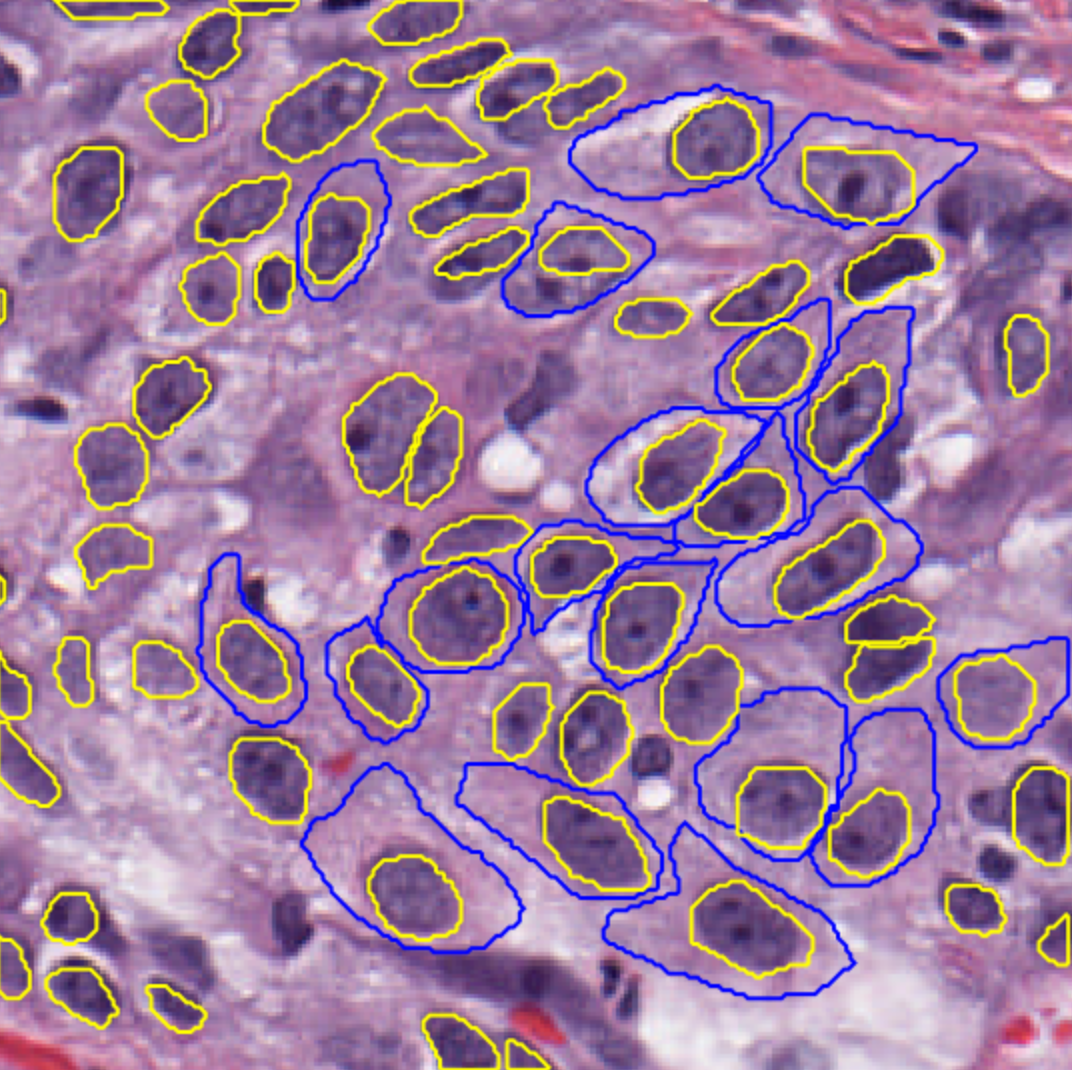}
    QuPath Finetuned
    \hspace{1.85cm}
    StarDist
    \hspace{1.85cm}
    Cellpose
    \hspace{1.85cm}
    Cyto R-CNN
    \hspace{1.85cm}
    Gold Standard
\caption{\label{fig:sample_prediction_results}Sample predictions on hematoxylin-eosin stained images of head and neck squamous cell carcinoma. All images are part of the test dataset. Note: Only tumor cells were annotated. Other cells (lymphocytes, macrophages, fibrocytes, etc.) were not included}
\end{figure*}

These performance differences can be well explained by looking at a few example predictions in figure \ref{fig:sample_prediction_results}. We can see sample predictions from each of the different models on a selected sample of images. The first column contains manual segmentations that are considered the gold standard for the purpose of this paper. 
The second column contains results from finetuned QuPath. Here we can see two things: First, not all nuclei are detected. Second, the cytoplasm shape is very uniform. This can be explained by the process of nucleus expansion: When expanding a given shape outwards by a fixed number of pixels, the resulting shape will appear more circular. Differences in the original shape's curvature will be smoothed out. This leads to all cells appearing somewhat similar.
The third column shows results of Cellpose. We can see that not all nuclei are found, which is particularly visible in the third row. Apart from that, we see that nucleus and cytoplasm do not have a one-to-one correspondence. Many nuclei are without cytoplasm and some cytoplasm predictions do not have an underlying nucleus. This is a direct result of how Cellpose was trained: One model has been trained for the nucleus and another model for the whole cell. This limitation of Cellpose does not seem to work well in our bright-field histological images. 
The fourth column shows results of our StarDist model. The nuclei segmentations are of high quality and are quite close to the gold standard. But the whole-cell segmentations are very uniform and always resemble a circular shape. Just like with QuPath, this is the result of nucleus expansion. StarDist itself is a nucleus-only segmentation model. Combining it with nucleus expansion is popular in the literature \citep{RN213, RN253}, but in our experiments does not yield optimal results.
In the last column, we see the results of Cyto R-CNN. The nuclei segmentations have a high quality and the cytoplasm shapes are much more diverse. While there are a few cases of multi-nuclei cells and overlapping segmentations, Cyto R-CNN is the only model in this comparison that is able to segment non-trivial nuclei shapes. 

\subsection{Segmentation accuracy}

Our segmentation results as reported in table \ref{table:performance_results} meet expectations from the literature. The watershed algorithm is a traditional tool that has been surpassed by application-specific neural networks \citep{RN78}. Consequently, it is no surprise that among the compared methods QuPath delivers the worst results for nucleus segmentation. It is important to note that tuning the watershed parameters has a large impact on the resulting performance (AP50 22.95\% compared to 35.24\%). With such a low accuracy for the nuclei, it is also no surprise that the cell segmentation as obtained by nucleus expansion is equally inaccurate (AP50 11.12\% and 19.46\%). 

Cellpose can already provide large improvements for both nucleus (AP50 48.35\%) and cell segmentation (AP50 31.85\%). However, Cellpose does not perform as well on our dataset as could be expected from its original publication \citep{RN135}. This is because our dataset contains exclusively HE-stained images, for which Cellpose was not originally intended. The color contrast in HE-stained images is much lower than in immunohistological images, for which Cellpose was originally built. The color gradient tracking algorithm which is used in Cellpose can thus not work as well. This is not a new observation. For example, \citet{RN223} reported similar results for Cellpose (AP50 46.30\%) when applying it to datasets of grayscale images. And \citet{RN225} reported non-optimal results Cellpose as well (AP50 64.40\%).

StarDist and Cyto R-CNN are the leading methods for nucleus segmentation in our experiments. StarDist achieves an AP50 of 70.36\%  while Cyto R-CNN achieves 78.32\%. The review from \citet{RN78} shows that these results are in line with expectations from the literature. In this review, the authors compare Mask R-CNN, watershed and a number of U-Net based architectures on the MoNuSeg dataset \citep{RN167}, which also contains HE images. Mask R-CNN achieves an AP50 of 78.59\%, which is similar to our results of Cyto R-CNN (78.32\%). The best U-Net architecture achieves an AP50 of 71.66\%, which is very similar to the results of StarDist in our experiments (70.36\%). Even though StarDist achieves great results in nucleus segmentation, combining it with nucleus expansion only results in suboptimal accuracies (AP50 45.33\%, AP75 2.32\%). Cyto R-CNN is able to significantly outperform that both in AP50 (58.65\%) and in AP75 (11.56\%). 

Our results for whole-cell segmentation are comparable to the results of \citet{RN228}. In this paper, the authors develop an approach for whole-cell segmentation in cytological images and evaluate its performance on two different public datasets. In cytological images, the cell nucleus and cell membrane are much more pronounced than in our HE-stained images. However, cells are often overlapping, which makes their task slightly different from ours. Nevertheless, the authors follow a similar approach when designing their architecture. They also extend Mask R-CNN and add some application-specific layers to it. Their new architecture outperforms all alternatives in their experiments and achieves a mean average precision of 64.02\% on one dataset and 49.43\% on another. The results of our Cyto R-CNN (AP50 58.65\%) are comparable to that. This shows that Mask R-CNN offers great potential for complex tasks such as whole-cell segmentation, because its two-stage design allows for application-specific extensions.

\subsection{Dataset}

To the authors' knowledge, CytoNuke is the first publicly available dataset of HE stained images which includes annotations for both the nucleus and the cytoplasm. Popular datasets such as PanNuke \citep{RN157} and MoNuSeg \citep{RN167} are limited to nuclei only. Thus, we decided to create our own quality-controlled dataset.
Generally, it is not easily possible to extend existing nuclei datasets with whole-cell annotations. PanNuke and MoNuSeg contain a number of different cell types (tumor, inflammatory, stromal cells) each of which have different cytoplasm characteristics. For some cells, the cytoplasm might not be visible at all or almost indistinguishable from the nucleus. This is often the case for lymphocytes for example. 
When creating the dataset, we intentionally limited it to one specific tumor type. Head and neck squamous cell carcinoma (HNSCC) was chosen for multiple reasons. First, its similarity to regular epithelium allows for a clear visual distinction of the cell membrane in HE stained images \citep{RN60}. Second, its cancer cells are known to be morphologically heterogeneous \citep{RN196, RN147}, making whole-cell segmentation a challenging task. Third, there are multiple hypotheses around the morphological-clinical connection in HNSCC that could benefit from accurate segmentation methods \citep{RN183, RN181, RN184, RN177, RN214}.
With 6,598 annotations (3,991 nuclei and 2,607 whole-cell), the size of our dataset is appropriate for a task as specific as whole-cell segmentation in bright-field histology. That is very comparable to the PanNuke dataset which contains around 3,000 nuclei annotations for HNSCC.

\subsection{Clinical impact}
There is a huge potential for extracting clinically important information from cellular measurements in bright-field whole slide images. There are a number of papers that investigate the relationship between cell morphology and clinical endpoints. For example, it has been shown that the PDL-1 score of non-small cell lung carcinoma can be accurately predicted using only HE-staining \citep{RN214}. And it was even found that certain nucleus shapes are associated with improved survival rates in HNSCC \citep{RN209}.

In absence of better methods, several papers make use of watershed, nucleus expansion or standard settings in QuPath. For example, \citet{RN253} used nucleus expansion in 2022 to analyze macrophages in Covid-19. \citet{RN213} used nucleus expansion to determine the presence of biomarkers in a cell. And \citet{RN211} used it in 2023 to study the tumor microenvironment of HNSCC. Using nucleus expansion in these cases can be problematic. As we have shown in this paper, nucleus expansion is generally not a reliable method to obtain cell segmentations.

Based on our results, we can derive the following recommendations for clinical analyses of cell segmentations in HE images: First, one should use a more sophisticated method than watershed. Neural networks such as Cellpose, StarDist and Cyto R-CNN all offer better performance. The precise choice will depend on the application and should be validated before-hand. Second, one should be careful when using cell expansion. There might be cases in which it can be appropriately used. But without prior knowledge about the cell, one should not assume a circular cell shape, which will result from nucleus expansion.

\subsection{Future work}

Despite these promising results, there are some clear limitations of our study. Most importantly, we have limited ourselves to a small dataset of one particular cell type. Head and neck squamous cell carcinoma was a great starting point for developing our method because of the clearly visible cell cytoplasm and heterogeneous morphology. It remains to be shown if the results of Cyto R-CNN can be replicated on larger datasets of different cell types. Our developed dataset will contribute to this, but also highlights the need for more comprehensive whole-cell datasets.

There is also potential for further refinement of our architecture. The cell scaling factor as described in section \ref{section:cytorcnn} was set to a constant value in our model. It could be possible to train the cell scaling factor as well. This would also make it easier to generalize our architecture to different cell types. 

In addition to that, Mask R-CNN was intentionally chosen as the base structure of our architecture, because its two-stage design allowed for easy extensibility. It might be possible to develop even better methods in the future, which make use of more recent advancements in computer vision, such as transformers for nuclei segmentation \cite{RN234}.

\section{Conclusion}

In this study, we have developed a new method to accurately segment the whole cell together with its nucleus in bright-field histological images. This method has been able to outperform all alternatives, including the popular setup of combining StarDist with nucleus expansion. Moreover, we found that existing methods can result in misleading data for cell measurements and should not be used for cytometric analysis without further validation. Our new method is able to improve the reliability of such morphological measurements, which could be used for future studies.

\section*{CRediT authorship contribution statement}
\textbf{Johannes Raufeisen:} Conceptualization, Methodology, Software, Validation, Formal analysis, Investigation, Data curation, Writing - Original Draft, Writing - Review \& Editing, Visualization;  \textbf{Kunpeng Xie:} Data curation; Validation; Writing - Review \& Editing; \textbf{Fabian Hörst:} Formal analysis; Writing -Review \& Editing; \textbf{Till Braunschweig:} Validation; Writing - Review \& Editing; \textbf{Jianning Li:} Formal analysis; Writing - Review \& Editing; \textbf{Jens Kleesiek:} Methodology; Writing - Review \& Editing Draft; \textbf{Jan Egger:} Formal analysis; Methodology; Writing - Review \& Editing; \textbf{Rainer Röhrig:} Formal analysis, Writing - Review \& Editing; \textbf{Bastian Leibe:} Resources, Writing - Review \& Editing, Supervision; \textbf{Frank Hölzle:} Resources, Writing - Review \& Editing, Supervision; \textbf{Alexander Hermans:} Conceptualization, Methodology, Writing - Review \& Editing, Supervision; \textbf{Behrus Puladi:} Conceptualization, Validation, Formal analysis, Investigation, Methodology, Resources, Data Curation, Writing - Review \& Editing, Visualization, Supervision, Project administration, Funding acquisition;

All authors have read and agreed to the published version of the manuscript.

\section*{Declaration of competing interest}
The authors declare no conflicts of interest.
 
\section*{Funding}
Behrus Puladi was funded by the Medical Faculty of RWTH Aachen University as part of the Clinician Scientist Program. We acknowledge FWF enFaced 2.0 [KLI 1044, \url{https://enfaced2.ikim.nrw/}] and KITE (Plattform für KI-Translation Essen) from the REACT-EU initiative [\url{https://kite.ikim.nrw/}, EFRE-0801977]. Fabian Hörst, Jianning Li, Jens Kleesiek and Jan Egger received funding from the Cancer Research Center Cologne Essen (CCCE).

\section*{Institutional Review Board Statement}
All experiments have been approved by the ethics commission of RWTH Aachen University under the project number EK 486/20.
 
\section*{Code Availability Statement}
The Cyto R-CNN code can be downloaded from \url{https://github.com/OMFSdigital/Cyto-R-CNN}.

\section*{Data Availability Statement}
The self-developed data set can be downloaded from Zenodo.org (\url{https://doi.org/10.5281/zenodo.10560728         }).

\section*{Acknowledgments}
Data used in this publication were generated by the National Cancer Institute Clinical Proteomic Tumor Analysis Consortium (CPTAC). Simulations were performed with computing resources granted by RWTH Aachen University under project rwth1193.

\normalsize
\vspace{+4mm}
\bibliography{refs}

\clearpage
\footnotesize
\onecolumn
\appendix

\begin{landscape}
\section*{Supplementary Material}

\begin{table}[H]
\caption{\label{table:cell_features_median_mean_iqr} Cell measurements}

\setlength{\tabcolsep}{0.5em} 
{\renewcommand{\arraystretch}{1.2}

\resizebox{23cm}{!}{%
\begin{tabular}{|l|lll|lll|lll|lll|lll|lll|}

\hline
\multirow{2}{*}{Feature} & \multicolumn{3}{l|}{Default QuPath}                               & \multicolumn{3}{l|}{Finetuned QuPath}                             & \multicolumn{3}{l|}{StarDist}                                     & \multicolumn{3}{l|}{Cellpose}                                     & \multicolumn{3}{l|}{Cyto R-CNN}                                    & \multicolumn{3}{l|}{Ground Truth}                                  \\ \cline{2-19} 
                         & \multicolumn{1}{l|}{Mean}   & \multicolumn{1}{l|}{Median} & IQR   & \multicolumn{1}{l|}{Mean}   & \multicolumn{1}{l|}{Median} & IQR   & \multicolumn{1}{l|}{Mean}   & \multicolumn{1}{l|}{Median} & IQR   & \multicolumn{1}{l|}{Mean}   & \multicolumn{1}{l|}{Median} & IQR   & \multicolumn{1}{l|}{Mean}   & \multicolumn{1}{l|}{Median} & IQR    & \multicolumn{1}{l|}{Mean}   & \multicolumn{1}{l|}{Median} & IQR    \\ \hline
Area, $\mu m^2$                     & \multicolumn{1}{l|}{104.13} & \multicolumn{1}{l|}{91.47}  & 68.93 & \multicolumn{1}{l|}{167.25} & \multicolumn{1}{l|}{158.39} & 78.00 & \multicolumn{1}{l|}{160.58} & \multicolumn{1}{l|}{149.34} & 61.79 & \multicolumn{1}{l|}{159.45} & \multicolumn{1}{l|}{146.91} & 89.50 & \multicolumn{1}{l|}{162.44} & \multicolumn{1}{l|}{147.40} & 102.34 & \multicolumn{1}{l|}{229.37} & \multicolumn{1}{l|}{202.10} & 229.37 \\ \hline
Perimeter, $\mu m$                & \multicolumn{1}{l|}{40.72}  & \multicolumn{1}{l|}{39.73}  & 63.87 & \multicolumn{1}{l|}{50.08}  & \multicolumn{1}{l|}{48.98}  & 12.33 & \multicolumn{1}{l|}{49.29}  & \multicolumn{1}{l|}{48.40}  & 9.30  & \multicolumn{1}{l|}{49.98}  & \multicolumn{1}{l|}{49.43}  & 14.12 & \multicolumn{1}{l|}{51.92}  & \multicolumn{1}{l|}{51.90}  & 18.73  & \multicolumn{1}{l|}{63.87}  & \multicolumn{1}{l|}{62.21}  & 22.15  \\ \hline
Circularity              & \multicolumn{1}{l|}{0.74}   & \multicolumn{1}{l|}{0.75}   & 0.66  & \multicolumn{1}{l|}{0.81}   & \multicolumn{1}{l|}{0.82}   & 0.09  & \multicolumn{1}{l|}{0.81}   & \multicolumn{1}{l|}{0.82}   & 0.08  & \multicolumn{1}{l|}{0.77}   & \multicolumn{1}{l|}{0.79}   & 0.10  & \multicolumn{1}{l|}{0.71}   & \multicolumn{1}{l|}{0.73}   & 0.12   & \multicolumn{1}{l|}{0.66}   & \multicolumn{1}{l|}{0.68}   & 0.15   \\ \hline
Solidity                 & \multicolumn{1}{l|}{0.94}   & \multicolumn{1}{l|}{0.94}   & 0.93  & \multicolumn{1}{l|}{0.97}   & \multicolumn{1}{l|}{0.97}   & 0.02  & \multicolumn{1}{l|}{0.97}   & \multicolumn{1}{l|}{0.98}   & 0.03  & \multicolumn{1}{l|}{0.96}   & \multicolumn{1}{l|}{0.97}   & 0.02  & \multicolumn{1}{l|}{0.95}   & \multicolumn{1}{l|}{0.96}   & 0.02   & \multicolumn{1}{l|}{0.93}   & \multicolumn{1}{l|}{0.94}   & 0.04   \\ \hline
Nucleus-to-cell ratio    & \multicolumn{1}{l|}{0.32}   & \multicolumn{1}{l|}{0.32}   & 0.12  & \multicolumn{1}{l|}{0.42}   & \multicolumn{1}{l|}{0.42}   & 0.08  & \multicolumn{1}{l|}{0.49}   & \multicolumn{1}{l|}{0.39}   & 0.10  & \multicolumn{1}{l|}{0.48}   & \multicolumn{1}{l|}{0.48}   & 0.29  & \multicolumn{1}{l|}{0.36}   & \multicolumn{1}{l|}{0.35}   & 0.17   & \multicolumn{1}{l|}{0.32}   & \multicolumn{1}{l|}{0.30}   & 0.22   \\ \hline
Max. diameter, $\mu m$            & \multicolumn{1}{l|}{14.55}  & \multicolumn{1}{l|}{14.13}  & 24.18 & \multicolumn{1}{l|}{17.58}  & \multicolumn{1}{l|}{17.00}  & 4.65  & \multicolumn{1}{l|}{17.52}  & \multicolumn{1}{l|}{17.30}  & 3.30  & \multicolumn{1}{l|}{18.04}  & \multicolumn{1}{l|}{17.55}  & 5.48  & \multicolumn{1}{l|}{18.77}  & \multicolumn{1}{l|}{18.63}  & 6.60   & \multicolumn{1}{l|}{24.18}  & \multicolumn{1}{l|}{23.16}  & 9.37   \\ \hline
Min. diameter, $\mu m$            & \multicolumn{1}{l|}{9.62}   & \multicolumn{1}{l|}{9.39}   & 13.11 & \multicolumn{1}{l|}{12.46}  & \multicolumn{1}{l|}{12.36}  & 2.86  & \multicolumn{1}{l|}{12.09}  & \multicolumn{1}{l|}{11.86}  & 3.14  & \multicolumn{1}{l|}{11.64}  & \multicolumn{1}{l|}{11.37}  & 4.36  & \multicolumn{1}{l|}{11.33}  & \multicolumn{1}{l|}{10.79}  & 4.54   & \multicolumn{1}{l|}{13.11}  & \multicolumn{1}{l|}{12.54}  & 5.51   \\ \hline
Hematoxylin median       & \multicolumn{1}{l|}{0.28}   & \multicolumn{1}{l|}{0.26}   & 0.16  & \multicolumn{1}{l|}{0.34}   & \multicolumn{1}{l|}{0.32}   & 0.19  & \multicolumn{1}{l|}{0.33}   & \multicolumn{1}{l|}{0.30}   & 0.18  & \multicolumn{1}{l|}{0.37}   & \multicolumn{1}{l|}{0.34}   & 0.20  & \multicolumn{1}{l|}{0.33}   & \multicolumn{1}{l|}{0.30}   & 0.19   & \multicolumn{1}{l|}{0.29}   & \multicolumn{1}{l|}{0.28}   & 0.14   \\ \hline
Hematoxylin mean         & \multicolumn{1}{l|}{0.31}   & \multicolumn{1}{l|}{0.29}   & 0.33  & \multicolumn{1}{l|}{0.38}   & \multicolumn{1}{l|}{0.38}   & 0.21  & \multicolumn{1}{l|}{0.37}   & \multicolumn{1}{l|}{0.37}   & 0.21  & \multicolumn{1}{l|}{0.40}   & \multicolumn{1}{l|}{0.39}   & 0.23  & \multicolumn{1}{l|}{0.36}   & \multicolumn{1}{l|}{0.35}   & 0.22   & \multicolumn{1}{l|}{0.33}   & \multicolumn{1}{l|}{0.31}   & 0.19   \\ \hline
Hematoxylin std. dev     & \multicolumn{1}{l|}{0.16}   & \multicolumn{1}{l|}{0.14}   & 0.16  & \multicolumn{1}{l|}{0.19}   & \multicolumn{1}{l|}{0.19}   & 0.13  & \multicolumn{1}{l|}{0.18}   & \multicolumn{1}{l|}{0.17}   & 0.12  & \multicolumn{1}{l|}{0.16}   & \multicolumn{1}{l|}{0.16}   & 0.11  & \multicolumn{1}{l|}{0.17}   & \multicolumn{1}{l|}{0.16}   & 0.12   & \multicolumn{1}{l|}{0.16}   & \multicolumn{1}{l|}{0.14}   & 0.11   \\ \hline
Hematoxylin max          & \multicolumn{1}{l|}{0.79}   & \multicolumn{1}{l|}{0.69}   & 0.92  & \multicolumn{1}{l|}{1.00}   & \multicolumn{1}{l|}{0.98}   & 0.64  & \multicolumn{1}{l|}{0.96}   & \multicolumn{1}{l|}{0.93}   & 0.63  & \multicolumn{1}{l|}{0.96}   & \multicolumn{1}{l|}{0.97}   & 0.62  & \multicolumn{1}{l|}{0.91}   & \multicolumn{1}{l|}{0.87}   & 0.63   & \multicolumn{1}{l|}{0.92}   & \multicolumn{1}{l|}{0.88}   & 0.62   \\ \hline
Hematoxylin min          & \multicolumn{1}{l|}{0.03}   & \multicolumn{1}{l|}{0.04}   & 0.05  & \multicolumn{1}{l|}{0.04}   & \multicolumn{1}{l|}{0.05}   & 0.10  & \multicolumn{1}{l|}{0.05}   & \multicolumn{1}{l|}{0.05}   & 0.10  & \multicolumn{1}{l|}{0.10}   & \multicolumn{1}{l|}{0.10}   & 0.11  & \multicolumn{1}{l|}{0.06}   & \multicolumn{1}{l|}{0.06}   & 0.11   & \multicolumn{1}{l|}{0.05}   & \multicolumn{1}{l|}{0.07}   & 0.11   \\ \hline
Eosin median             & \multicolumn{1}{l|}{0.22}   & \multicolumn{1}{l|}{0.27}   & 0.20  & \multicolumn{1}{l|}{0.20}   & \multicolumn{1}{l|}{0.22}   & 0.22  & \multicolumn{1}{l|}{0.20}   & \multicolumn{1}{l|}{0.19}   & 0.21  & \multicolumn{1}{l|}{0.20}   & \multicolumn{1}{l|}{0.20}   & 0.21  & \multicolumn{1}{l|}{0.21}   & \multicolumn{1}{l|}{0.26}   & 0.22   & \multicolumn{1}{l|}{0.23}   & \multicolumn{1}{l|}{0.27}   & 0.18   \\ \hline
Eosin mean               & \multicolumn{1}{l|}{0.22}   & \multicolumn{1}{l|}{0.27}   & 0.22  & \multicolumn{1}{l|}{0.20}   & \multicolumn{1}{l|}{0.21}   & 0.22  & \multicolumn{1}{l|}{0.20}   & \multicolumn{1}{l|}{0.17}   & 0.22  & \multicolumn{1}{l|}{0.20}   & \multicolumn{1}{l|}{0.17}   & 0.22  & \multicolumn{1}{l|}{0.21}   & \multicolumn{1}{l|}{0.25}   & 0.22   & \multicolumn{1}{l|}{0.22}   & \multicolumn{1}{l|}{0.27}   & 0.19   \\ \hline
Eosin std. dev           & \multicolumn{1}{l|}{0.07}   & \multicolumn{1}{l|}{0.07}   & 0.07  & \multicolumn{1}{l|}{0.08}   & \multicolumn{1}{l|}{0.08}   & 0.04  & \multicolumn{1}{l|}{0.08}   & \multicolumn{1}{l|}{0.07}   & 0.04  & \multicolumn{1}{l|}{0.08}   & \multicolumn{1}{l|}{0.07}   & 0.04  & \multicolumn{1}{l|}{0.07}   & \multicolumn{1}{l|}{0.07}   & 0.04   & \multicolumn{1}{l|}{0.07}   & \multicolumn{1}{l|}{0.06}   & 0.04   \\ \hline
Eosin max                & \multicolumn{1}{l|}{0.43}   & \multicolumn{1}{l|}{0.42}   & 0.42  & \multicolumn{1}{l|}{0.43}   & \multicolumn{1}{l|}{0.42}   & 0.20  & \multicolumn{1}{l|}{0.42}   & \multicolumn{1}{l|}{0.41}   & 0.20  & \multicolumn{1}{l|}{0.42}   & \multicolumn{1}{l|}{0.42}   & 0.22  & \multicolumn{1}{l|}{0.41}   & \multicolumn{1}{l|}{0.41}   & 0.21   & \multicolumn{1}{l|}{0.42}   & \multicolumn{1}{l|}{0.42}   & 0.22   \\ \hline
Eosin min                & \multicolumn{1}{l|}{-0.01}  & \multicolumn{1}{l|}{0.05}   & -0.05 & \multicolumn{1}{l|}{-0.09}  & \multicolumn{1}{l|}{-0.09}  & 0.39  & \multicolumn{1}{l|}{-0.08}  & \multicolumn{1}{l|}{-0.08}  & 0.39  & \multicolumn{1}{l|}{-0.06}  & \multicolumn{1}{l|}{-0.06}  & 0.39  & \multicolumn{1}{l|}{-0.05}  & \multicolumn{1}{l|}{-0.05}  & 0.38   & \multicolumn{1}{l|}{-0.05}  & \multicolumn{1}{l|}{-0.04}  & 0.37   \\ \hline

\end{tabular}%
} 
} 
\end{table}

\end{landscape}
\end{document}